# Bluish Veil Detection and Lesion Classification using Custom Deep Learnable Layers with Explainable Artificial Intelligence (XAI)

**M. A. Rasel[1], Sameem Abdul Kareem[1], Zhenli Kwan[2], Shin Shen Yong[2], Unaizah Obaidellah[1,*]**

[1]Department of Artificial Intelligence, Faculty of Computer Science and Information Technology, Universiti Malaya, Kuala Lumpur, 50603, Malaysia

[2]Division of Dermatology, Department of Medicine, Faculty of Medicine, Universiti Malaya, Kuala Lumpur, 50603, Malaysia

[*]Corresponding Author: Unaizah Obaidellah. Email: unaizah@um.edu.my

**Abstract:** Melanoma, one of the deadliest types of skin cancer, accounts for thousands of fatalities globally. The bluish, blue-whitish, or blue-white veil (BWV) is a critical feature for diagnosing melanoma, yet research into detecting BWV in dermatological images is limited. This study utilizes a non-annotated skin lesion dataset, which is converted into an annotated dataset using a proposed imaging algorithm (color threshold techniques) on lesion patches based on color palettes. A Deep Convolutional Neural Network (DCNN) is designed and trained separately on three individual and combined dermoscopic datasets, using custom layers instead of standard activation function layers. The model is developed to categorize skin lesions based on the presence of BWV. The proposed DCNN demonstrates superior performance compared to the conventional BWV detection models across different datasets. The model achieves a testing accuracy of 85.71% on the augmented PH2 dataset, 95.00% on the augmented ISIC archive dataset, 95.05% on the combined augmented (PH2+ISIC archive) dataset, and 90.00% on the Derm7pt dataset. An explainable artificial intelligence (XAI) algorithm is subsequently applied to interpret the DCNN's decision-making process about the BWV detection. The proposed approach, coupled with XAI, significantly improves the detection of BWV in skin lesions, outperforming existing models and providing a robust tool for early melanoma diagnosis.

## 1. Introduction

Melanoma is one of the most aggressive and deadly forms of skin cancer, responsible for a significant number of deaths worldwide each year ***(Wróblewska-Łuczka et al., 2023)***. The incidence of melanoma has been increasing globally, emphasizing the need for early and accurate diagnosis to improve patient outcomes. Early detection is critical as melanoma is highly treatable in its initial stages and can quickly become life-threatening if it spreads to other parts of the body ***(Switzer et al., 2022)***.

Various clinical methods have been developed to aid in the early diagnosis of melanoma. These include the Three-Point Checklist ***(Soyer et al., 2004)***, the Seven-Point Checklist ***(Walter et al., 2013)***, and the CASH algorithm ***(Henning et al., 2008)***. These methods rely on a combination of criteria to examine skin lesions, such as asymmetry, border irregularity, color variation, diameter, and the presence of specific dermoscopic features like streaks, and dots-globules ***(Garrison et al., 2023)***. Among these criteria, the bluish, blue-whitish or blue-white veil (BWV) is particularly noteworthy. The BWV is characterized by a blue and structureless zone with an overlying white "ground-glass" haze, typically found in raised or palpable areas of a lesion ***(Seidenari et al., 2006)***. **Fig. 1** shows two skin lesions with BWV from the PH2 ***(Mendonca et al., 2013)*** dataset. It is a significant indicator of melanoma ***(Ciudad-Blanco et al., 2014)***, often observed in more advanced lesions, and its presence necessitates careful examination and prompt medical intervention.

Despite its clinical importance, detecting BWV can be challenging due to its subtle appearance. The veil is often formed of a faint mix of gray, blue, and white patch ***(De Giorgi et al., 2003; Madooei and Drew, 2013; Soyer et al., 2004; Walter et al., 2013)***, which can be easily overlooked against the complex background of other lesion colors. Accurate detection requires significant expertise and a keen eye, making it a task prone to human error, particularly in busy clinical settings.

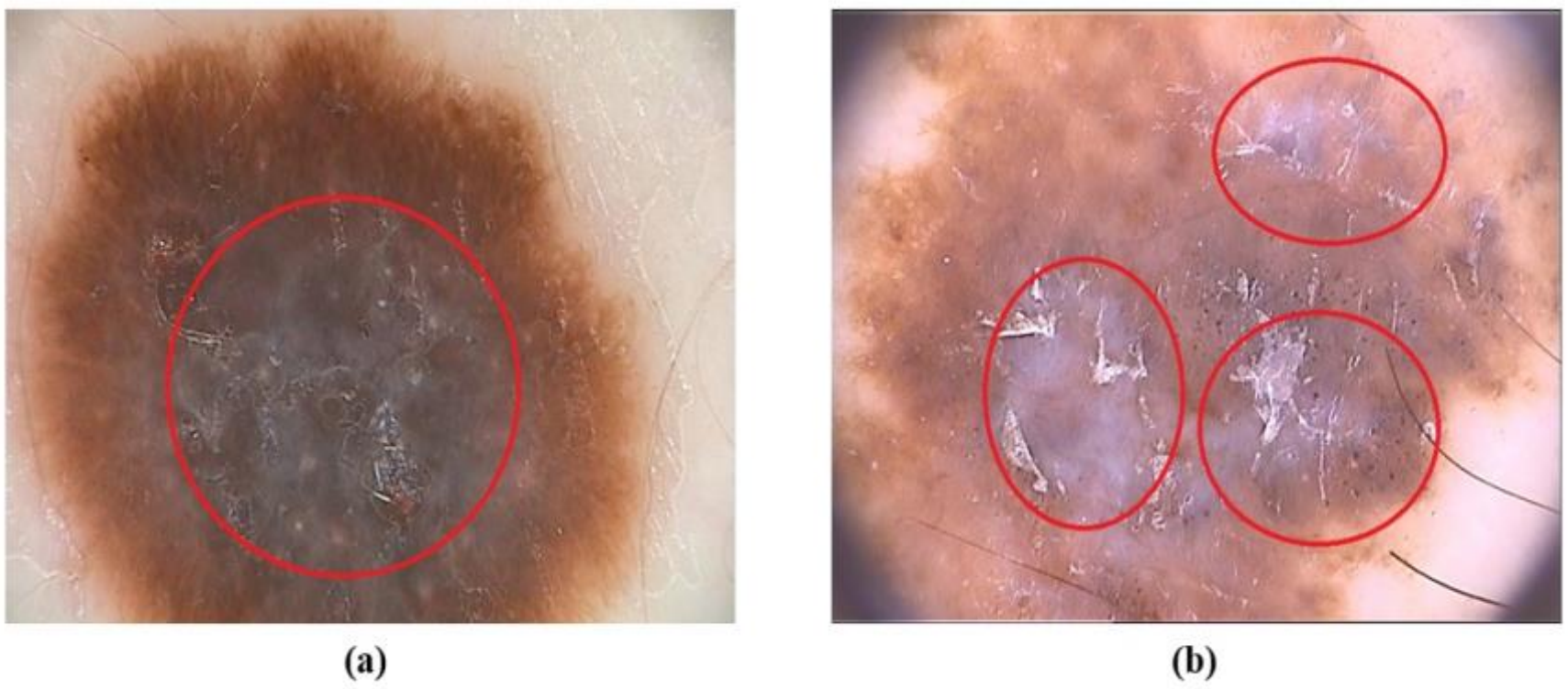

**Fig. 1:** The BWV (red circle/s) in the dermoscopic images (a) and (b) of the PH2 dataset ***(Mendonca et al., 2013)***.

Given the critical role of BWV in melanoma diagnosis, there is an urgent need for reliable and automated detection methods. Current manual examination techniques are not sufficiently robust, often leading to missed diagnoses of early-stage melanomas. Moreover, there is a lack of comprehensive research focusing on the automated detection of BWV using advanced technologies such as deep learning.

Deep Convolutional Neural Networks (DCNNs) have shown great promise in various image recognition tasks ***(Manakitsa et al., 2024)***, including medical imaging ***(Tsuneki, 2022; Olayah et al., 2023)***. However, their application to BWV detection in dermoscopic images has been limited. Existing methods do not adequately address the complexities involved in identifying BWV, and there is a clear gap in leveraging state-of-the-art machine learning techniques to enhance detection accuracy and reliability.

This study introduces a novel approach to the detection of BWV in dermoscopic images by leveraging deep convolutional neural networks (DCNNs) combined with explainable artificial intelligence (XAI) techniques. Unlike previous studies that rely on traditional image processing and manual inspection, our approach automates the detection process, enhancing accuracy and consistency. The use of XAI further provides insights into the model's decision-making process, ensuring transparency and trust in its predictions. This research fills a significant gap in existing literature by offering a comprehensive, automated solution for BWV detection, potentially improving early melanoma diagnosis.

The primary aim of this study is to develop a DCNN capable of accurately detecting BWV in skin lesion images. Specific objectives include:

i. Creating an annotated dataset of skin lesions using a proposed imaging algorithm based on color threshold techniques. This involves converting a non-annotated skin lesion dataset into an annotated one, facilitating the training of the DCNN.
ii. Designing and training a DCNN with custom layers optimized for BWV detection. This involves experimenting with different network architectures and hyperparameters to achieve optimal performance.
iii. Evaluating the proposed DCNN's performance on multiple dermoscopic datasets to assess its generalizability and robustness. This includes testing the model on unfamiliar datasets to verify its effectiveness in diverse clinical scenarios.
iv. Applying explainable artificial intelligence (XAI) techniques to interpret the DCNN's decision-making process.

This study focuses on the detection of BWV in dermoscopic images using a threshold-based imaging algorithm and a DCNN. The imaging algorithm, implemented for data annotation and validated by two clinical experts, is an essential component of this work. The study encompasses the development, training, and evaluation of the proposed methods across several datasets, aiming to provide a reliable tool to assist dermatologists and skin cancer experts in specifying the BWV characteristics. By improving early melanoma detection in dermoscopic images, this study contributes to more accurate and timely diagnoses.

This manuscript is organized as follows: Section 2 reviews related works in BWV detection on skin lesion images. Section 3 describes the methodology, including data acquisition, establishing ground truth, model design, and training procedures. Section 4 presents the experimental results and comparative analysis and discusses the implications of the findings and the proposed model's decision-making process. At the end, Section 5 concludes with future research directions.

## 2. Related Works and Research Gap

There are not many studies that report an approach, experimental procedure, and results specifically about the detection of the feature under study here. A handful of studies aim to detect (and localize) BWV in skin lesion images. Celebi et al. ***(2006)*** proposed a machine-learning approach to the detection of BWV areas in dermoscopic images. The presented approach was comprised of several steps including preprocessing, color feature extraction, decision tree induction, rule application, and post-processing. The results were evaluated visually by a dermatologist on a set of dermoscopic images and found to be satisfactory. Celebi et al. ***(2008)*** proposed another machine-learning approach to the detection of BWV in dermoscopic images. The method was also comprised of several steps including preprocessing, feature extraction, decision tree induction, rule application, and post-processing. The detected BWVs were characterized using a numerical feature, which in conjunction with an ellipticity measure yielded a sensitivity of 69.35% and a specificity of 89.97% on a set of dermoscopic images. Di Leo et al. ***(2009)*** proposed the automatic detection of the BWV and Regression in digital epiluminescence microscopy (ELM) images, which constitute respectively a major and a minor criterion for the estimation of the malignancy of a skin lesion according to a well-known diagnostic method. Fabbrocini et al. ***(2014)*** proposed an image processing setup that allowed the automatic detection of some specific dermoscopic criteria. They analyzed the BWV, the regression, and the irregular streaks. The procedure developed was tested by considering a set of ELM images. Arroyo et al. ***(2011)*** presented a complete stack of algorithms for the computer-aided detection of the BWV pattern, with the help of supervised machine learning techniques, and produced good results. Wadhawan et al. ***(2012)*** proposed an algorithm that had high accuracy for the detection of BWV and was invariant to light-intensity scaling and shifting. Even though the first global approach was slightly less accurate compared to the trivial global approach, it was much more stable due to the change in user-defined parameters. Madooei and Drew ***(2013)*** proposed two schemas for automatic detection of BWV feature in dermoscopic images. They first proposed a revised threshold-based method with results comparable to state-of-the-art, with much-reduced computation. The second approach, their main contribution, set out an innovative method that attempted to mimic human interpretation of lesion colors. The latter outperformed prior art and introduced a perceptually and semantically meaningful new approach that can serve as a scaffolding for new color investigations in dermoscopic, for example for detection and recognition of common colors under dermoscopic. Kropidlowski et al. ***(2016)*** presented the implementation of three criteria from the Seven-Point Checklist for diagnosing melanoma. Those were the presence of BWV, the atypical vascular pattern, and the regression structures. Application of proposed detection algorithms showed promising results which were slightly better results than the competitive approach. Madooei et al. ***(2019)*** proposed a new approach for automatic identification of the BWV feature which needs considerably less supervision than for previous methods. Their method employed the multiple instance learning (MIL) framework to learn from image-level labels, without explicit annotation of samples of image regions containing the feature under study. Cacciapuoti et al. ***(2020)*** proposed an image-based measurement system for the automatic detection of melanoma according to a well-known diagnostic method (including BWV detection) as support to the dermatologist activity. It adopted advanced statistical techniques to perform the main tasks (automatic recognition of the skin lesion within the dermoscopic image, measurement of morphological and chromatic parameters, detection of the dermoscopic structures included in the Seven-Point Checklist method, and overall classification of the lesions) necessary for providing a second opinion about the clinical decision. These research works are summarized from a technical point of view in **Table 1** to find the limitations and gaps in determining the objectives of this research.

**Table 1**
A summary of existing approaches for detecting BWV on skin lesion images.

| Author | Year | Feature Extraction | Dataset | No. of data | Classifier | Future Work | Limitation |
|---|---|---|---|---|---|---|---|
| Celebi et al. ***(2006)*** | 2006 | Color | Atlas | 224 | Decision Tree | Incorporating texture features | Limited data |
| Celebi et al. ***(2008)*** | 2008 | Color and texture | Atlas | 545 | Decision Tree | Not mentioned | Single dataset |
| Di Leo et al. ***(2009)*** | 2009 | Color | Atlas | 210 | Logistic Model Tree | Not mentioned | Limited data |
| Fabbrocini et al. ***(2010)*** | 2010 | Color | Atlas | 200 | Logistic Model Tree | Not mentioned | Limited data |
| Arroyo et al. ***(2011)*** | 2011 | Color | Own | 887 | Decision Tree | Integrating other patterns | Single dataset |
| Wadhawan et al. ***(2012)*** | 2012 | Color | Own | 1009 | Global level | Not mentioned | Single dataset |
| Madooei and Drew ***(2013)*** | 2013 | Color | - | - | Decision Tree | Implementing on other colors | No dataset |
| Kropidlowski et al. ***(2016)*** | 2016 | Color | Own | 193 | Manual | Improving current algorithm | Limited data |
| Madooei et al. ***(2019)*** | 2019 | Color and texture | Atlas, PH2 | 855 200 | MIL | Adapting multi-label MIL | Complex to implement |
| Cacciapuoti et al. ***(2020)*** | 2020 | Color | Own | 270 | Logistic Model Tree | Investigating other structures | Limited data |

In the realm of dermoscopy for melanoma detection, a significant research gap exists concerning the accurate and reliable detection of BWV in skin lesion images. Current approaches often struggle to precisely identify and differentiate BWV from other structures, leading to potential misclassifications and diagnostic errors. There is a need for a deep learning-based approach to distinguish between the presence and absence of BWV in lesions, providing a more accurate classification than existing BWV detection methods. Additionally, only a limited amount of image data was used to evaluate the various proposed BWV detection algorithms listed in **Table 1**. There is a requirement to use multiple sources (different datasets) to demonstrate the adaptability and viability of the BWV detection algorithm across various origins. Furthermore, an imaging technique is needed for analyzing skin lesions, which should help convert a non-annotated dataset into an annotated one based on the presence of BWV. The newly annotated dataset can then be used to provide sufficient data for training a deep learning model.

## 3. Methods

The input image data for BWV detection is gathered from three distinct dermoscopic datasets: PH2, ISIC archive, and Derm7pt. Augmentation is necessary for the PH2 and ISIC archive datasets due to a limited number of images. Among the three original datasets, two are annotated, while one remains non-annotated. To transform the non-annotated dataset into an annotated one, preprocessing is essential (a detailed discussion is available in the following sections). Subsequently, a deep learning model is devised and trained using these three datasets.

To comprehend the decision-making process of this trained deep learning model, explainable artificial intelligence is employed. Finally, the trained model is applied to detect skin lesions with BWV on unfamiliar data. **Fig. 2** illustrates the comprehensive process of the proposed BWV detection approach on skin lesion images.

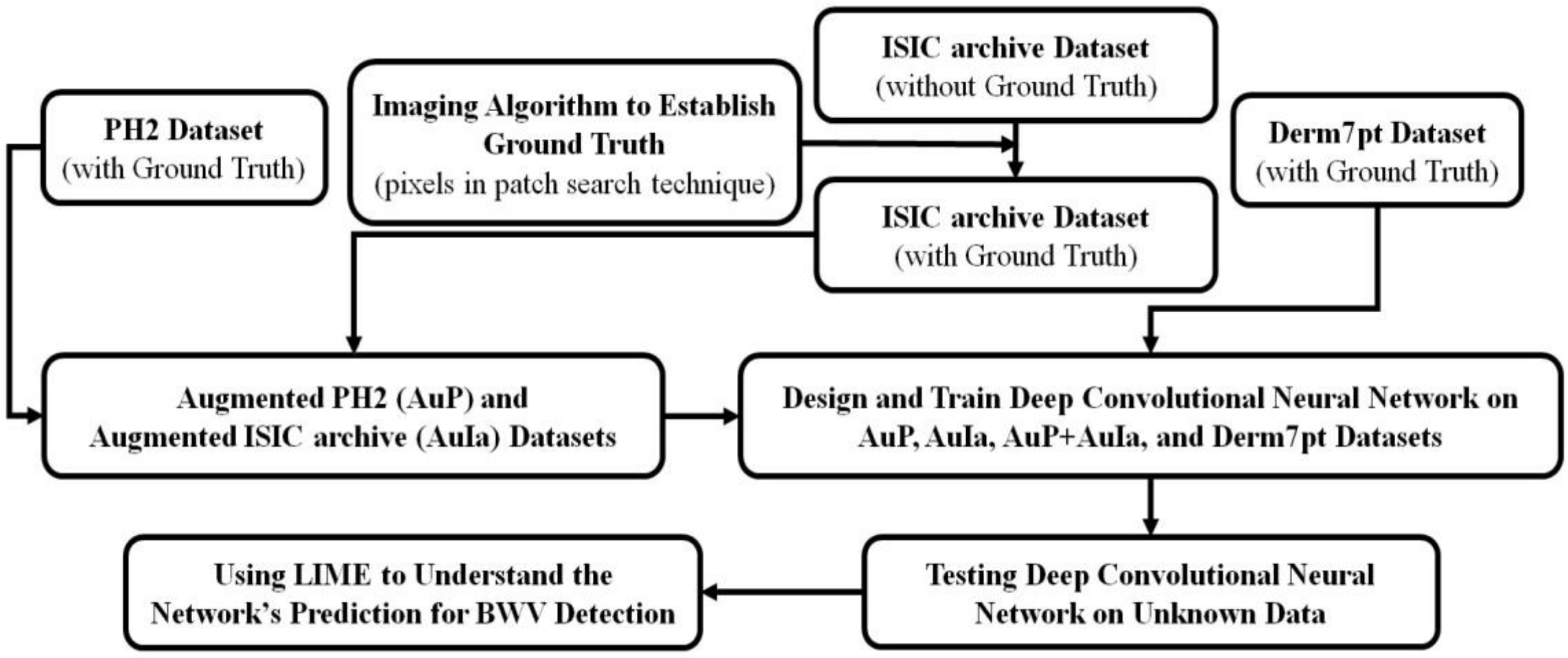

**Fig. 2:** The overall process of the proposed BWV detection approach on different datasets including data augmentation, model designing, model training, and model testing process.

### *3.1 Data Acquisition*

PH2 is a dermoscopic image database acquired at the Dermatology Service of Hospital Pedro Hispano, Matosinhos, Portugal ***(Mendonca et al., 2013)*** under the same conditions through the Tuebinger Mole Analyzer system using a magnification of 20x. They are 8-bit RGB color images with a resolution of 768x560 pixels. This image database contains a total of 200 dermoscopic images of melanocytic lesions, including 120 BWV, and 80 non-BWV.

An additional 204 images are obtained from the ISIC archive, encompassing datasets such as ISIC2016 ***(Gutman et al., 2016)***, ISIC2017 ***(Codella et al., 2017)***, ISIC2018 ***(Codella et al., 2018; Tschandl, 2018)***, and ISIC2019 ***(Tschandl, 2018; Codella et al., 2017; Combalia et al., 2019)***, where thousands of dermoscopic images depicting various skin diseases are available for research purposes. For the BWV detection experiment, 104 melanoma images are randomly selected from a pool of 5598, along with 100 benign images from 47,684. As these images lacked annotations, we utilized our dermatological and computer vision expertise to annotate them (the senior consultant dermatologist annotated the data separately, and discrepancies were discussed and resolved), providing information on BWV and non-BWV (see in section 3.3). These annotated data are used for evaluating the output of the proposed imaging algorithm. The images varied in size, ranging from 640x480 pixels to 1987x1987 pixels, with 66 images featuring BWV and 138 images non-BWV among the total 204 images.

Derm7pt is a database for evaluating computerized image-based prediction of the Seven-Point skin lesion malignancy checklist ***(Kawahara et al., 2019)***. The dataset includes 2013 clinical and dermoscopic color images, along with corresponding structured metadata suitable for training and evaluating computer-aided diagnosis systems. It contains a total of 381 BWV and 1632 non-BWV dermoscopic images.

### *3.2 Evaluation Metrics*

The evaluation metrics such as accuracy (*AC*), precision (*PR*), Sensitivity (*SE*), F1-score (*F1*), and specificity (*SP*) are applied to the confusion matrices of the experiments. The mathematical form of these evaluation metrics is-

$$AC = (TP + TN)/(P + N) \quad (1)$$

$$PR = TP/(TP + FP) \quad (2)$$

$$SE = TP/(TP + FN) \quad (3)$$

$$F1 = 2TP/(2TP + FP + FN) \quad (4)$$

$$SP = TN/(FP + TN) \quad (5)$$

Here, *TP* is true positive, *TN* is true negative, *P* is total positive, *N* is total negative, *FP* is false positive, and *FN* is false negative of the confusion matrix.

### *3.3 Establishing Ground Truth of BWV on ISIC Archive*

The ISIC archive dataset lacks ground truth information regarding the presence of BWV. Since BWV is not always easy to visually identify and recognize, a computerized automatic color analysis method can provide clinicians with an objective second opinion. To establish ground truth, 80 previously published color palettes are utilized ***(Madooei and Drew, 2013)***, selected based on the most frequently occurring colors associated with BWV, as depicted in **Fig. 3**. These palettes informed the determination of pixel values for each color, detailed in **Table 2** across RGB channels. These 80 color palettes were selected to mimic the human perception of BWV colors which are not directly a mix of blue and white colors, more likely the color of blurriness when veils cover the lesion ***(Madooei and Drew, 2013)***, ***(Landa and Fairchild, 2005)***. Analyzing this data, the minimum and maximum values for red, green, and blue channels are identified, which are respectively 45, 73, 73 (minimum) and 166, 98, 98 (maximum), constituting the considered color range for BWV.

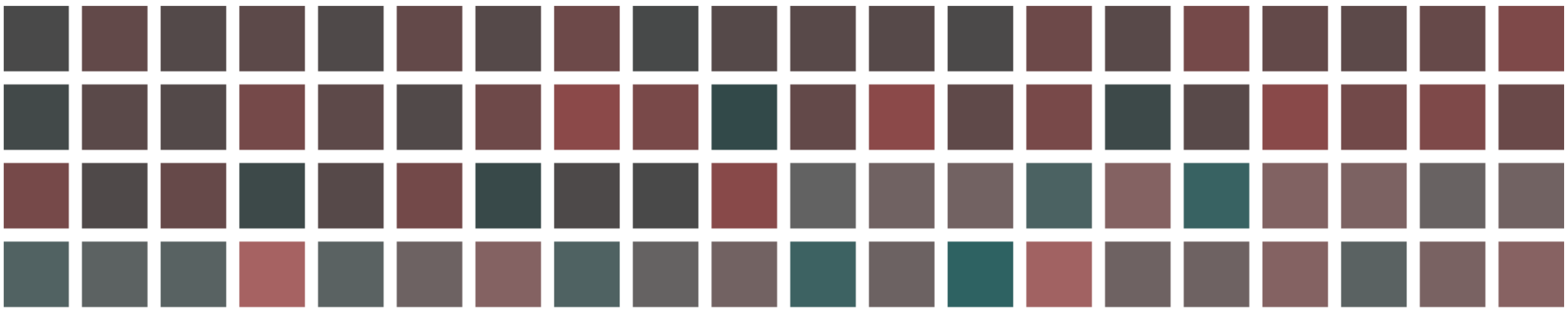

**Fig. 3:** The 80 color palettes based on the most frequent colors identified as BWV skin lesions ***(Madooei and Drew, 2013)***.

**Table 2**
RGB channels' values (0 to 255) for 80 color palettes from **Fig. 3**.

| **Channel** | **Values (0 to 255)** | | | | | | | | | | | | | | | | | | | |
|---|---|---|---|---|---|---|---|---|---|---|---|---|---|---|---|---|---|---|---|---|
| **R** | 73 | 98 | 83 | 92 | 79 | 97 | 85 | 108 | 71 | 82 | 80 | 86 | 75 | 109 | 89 | 119 | 96 | 94 | 103 | 125 |
| **G** | 73 | 73 | 73 | 73 | 73 | 73 | 73 | 73 | 73 | 73 | 73 | 73 | 73 | 73 | 73 | 73 | 73 | 73 | 73 | 73 |
| **B** | 73 | 73 | 73 | 73 | 73 | 73 | 73 | 73 | 73 | 73 | 73 | 73 | 73 | 73 | 73 | 73 | 73 | 73 | 73 | 73 |
| **R** | 66 | 90 | 84 | 117 | 93 | 81 | 110 | 138 | 121 | 50 | 99 | 139 | 95 | 120 | 62 | 88 | 137 | 114 | 126 | 106 |
| **G** | 73 | 73 | 73 | 73 | 73 | 73 | 73 | 73 | 73 | 73 | 73 | 73 | 73 | 73 | 73 | 73 | 73 | 73 | 73 | 73 |
| **B** | 73 | 73 | 73 | 73 | 73 | 73 | 73 | 73 | 73 | 73 | 73 | 73 | 73 | 73 | 73 | 73 | 73 | 73 | 73 | 73 |
| **R** | 118 | 78 | 102 | 61 | 87 | 115 | 56 | 77 | 74 | 136 | 98 | 112 | 116 | 75 | 130 | 56 | 129 | 124 | 104 | 113 |
| **G** | 73 | 73 | 73 | 73 | 73 | 73 | 73 | 73 | 73 | 73 | 98 | 98 | 98 | 98 | 98 | 98 | 98 | 98 | 98 | 98 |
| **B** | 73 | 73 | 73 | 73 | 73 | 73 | 73 | 73 | 73 | 73 | 98 | 98 | 98 | 98 | 98 | 98 | 98 | 98 | 98 | 98 |
| **R** | 81 | 92 | 88 | 166 | 90 | 109 | 131 | 79 | 101 | 114 | 61 | 108 | 46 | 161 | 110 | 111 | 132 | 91 | 121 | 135 |
| **G** | 98 | 98 | 98 | 98 | 98 | 98 | 98 | 98 | 98 | 98 | 98 | 98 | 98 | 98 | 98 | 98 | 98 | 98 | 98 | 98 |
| **B** | 98 | 98 | 98 | 98 | 98 | 98 | 98 | 98 | 98 | 98 | 98 | 98 | 98 | 98 | 98 | 98 | 98 | 98 | 98 | 98 |

RGB is a standard and widely used color model, especially in digital imaging and display technologies. Most digital cameras capture images in the RGB color space. The ISIC archive is a collection of RGB images that were captured under the constant condition (also normalized to make sure constant spectrum). Using RGB simplifies interoperability with various image processing and computer vision algorithms which are conceptually simple and intuitive. This model directly corresponds to the primary colors of light (red, green, and blue), which are fundamental to human vision. This simplicity makes RGB easy to understand and work with for BWV detection. On the other side, the input layer for lesion image classification of the proposed deep learning model (second part of this research) is defined with a shape corresponding to the image dimensions, typically expressed as (height, width, and channels), where channels represent the three-color channels (Red, Green, Blue).

A systematic analysis (including small patch-based and pixel-based) is conducted of each image within the ISIC archive dataset to identify BWV. **Fig. 4** shows a dermoscopic image of the ISIC archive dataset. Initial image segmentation involves dividing each image into 16x16 pixel patches. An example is shown in **Fig. 5**. These patches are individually scrutinized to identify colors within the established RGB channel range for BWV from **Table 2**. To mitigate noisy outputs, the analysis is performed patch-by-patch, considering a full patch as BWV if any single patch contained BWV color. **Fig. 6** shows the red marks on those patches which exactly matched with the RGB channel range for BWV. If an image contains at least one patch resembling BWV color, it is considered indicative of BWV and potentially melanoma.

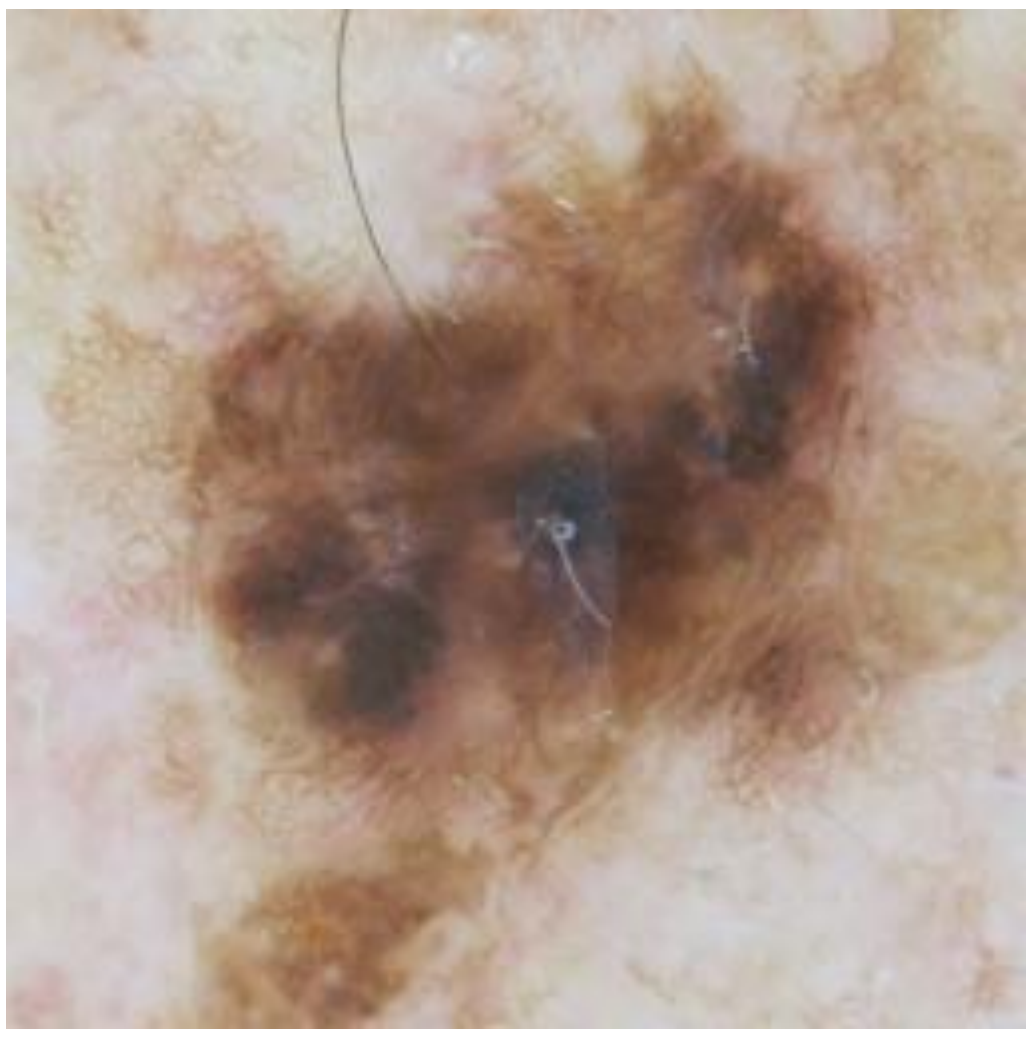

**Fig. 4:** A dermoscopic image (ISIC_0072405.jpg) from ISIC archive: ISIC2016 ***(Gutman et al., 2016)***, ISIC2017 ***(Codella et al., 2017)***, ISIC2018 ***(Codella et al., 2018; Tschandl, 2018)***, and ISIC2019 ***(Tschandl, 2018; Codella et al., 2017; Combalia et al., 2019)***.

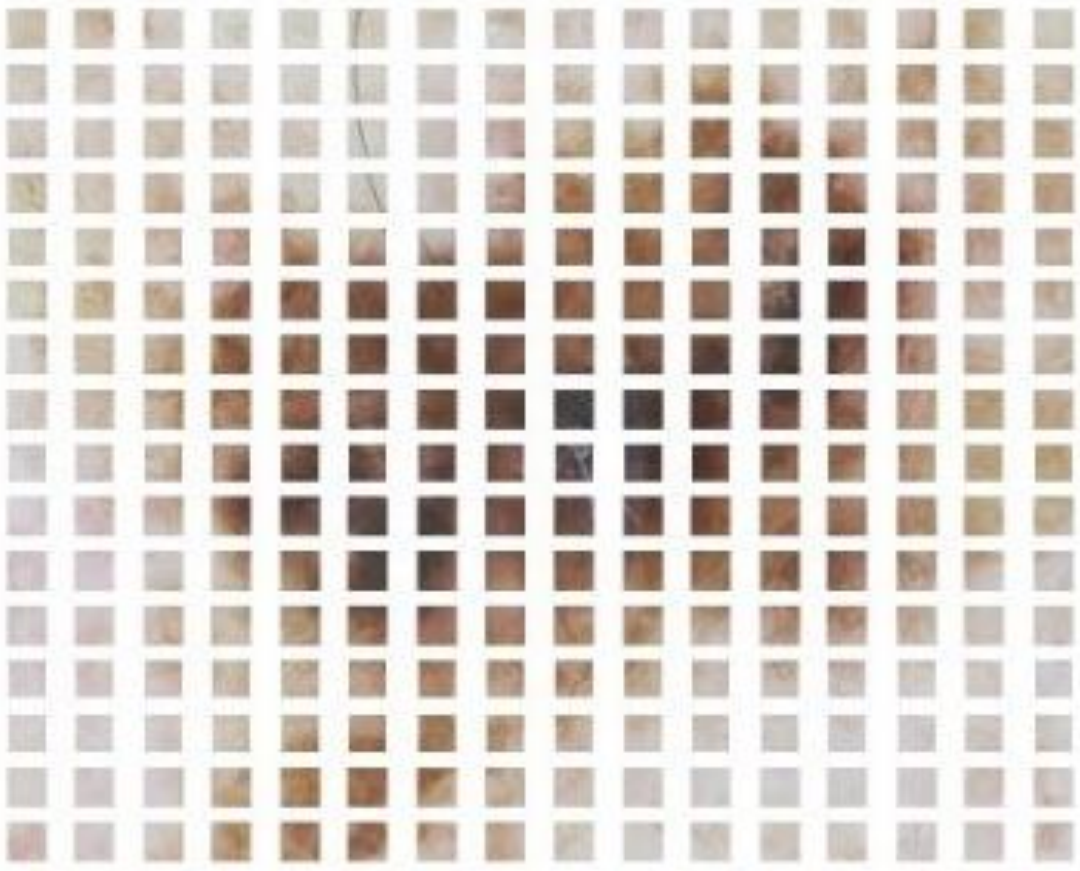

**Fig. 5:** The dermoscopic image is divided into small patches (16x16 pixels).

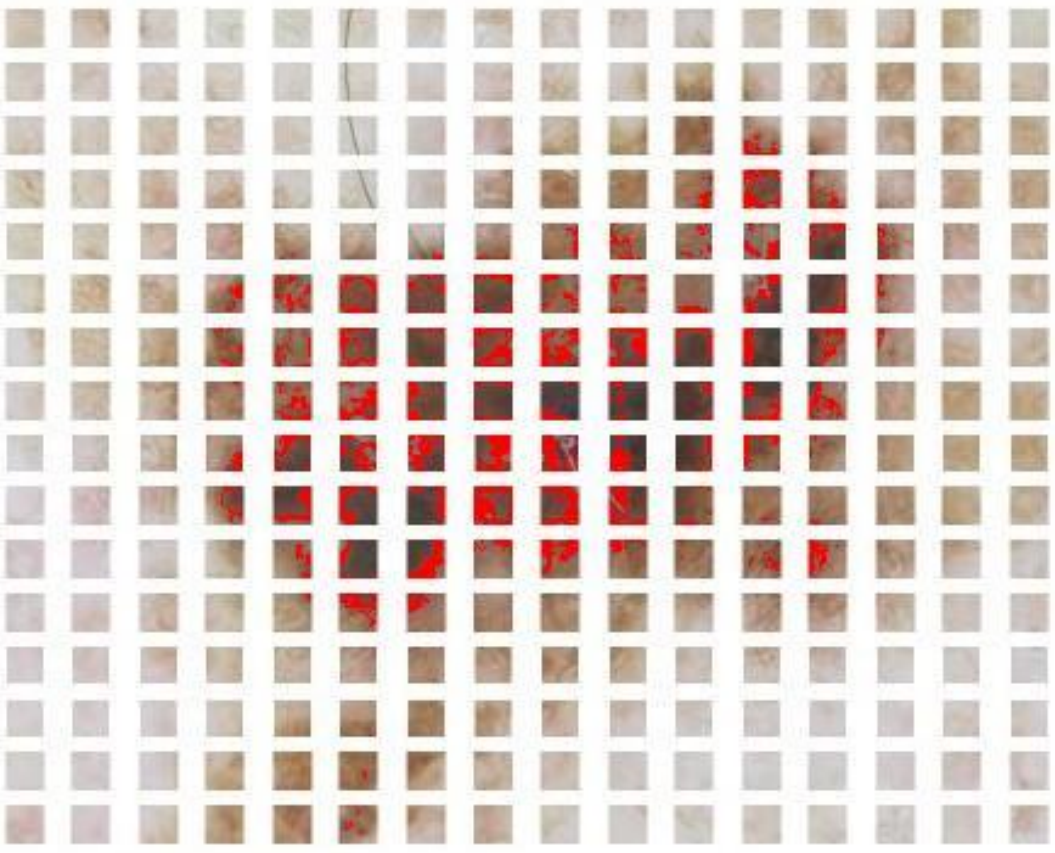

**Fig. 6:** Red marks on the patches indicate the presence of BWV.

**Table 3** describes the imaging algorithm (pixel search technique on the patches) by visualizing the presence of BWV on the input image. Finding those specific patches in the lesion images involved locating the pixel's coordinates (row and column) within the image's grid of pixels. This is typically done using the image's (x, y) coordinates, where 'x' represents the column and 'y' represents the row.

**Table 3**
Description of imaging algorithm on the lesion images of PH2 ***(Mendonca et al., 2013)*** and ISIC archive: ISIC2016 ***(Gutman et al., 2016)***, ISIC2017 ***(Codella et al., 2017)***, ISIC2018 ***(Codella et al., 2018; Tschandl, 2018)***, and ISIC2019 ***(Tschandl, 2018; Codella et al., 2017; Combalia et al., 2019)***.

| Dataset/ Image ID | Original | Presence of BWV | Processed Image | Lesion Patches(16x16) | RGB Value | |
|---|---|---|---|---|---|---|
| PH2 IMD031 | 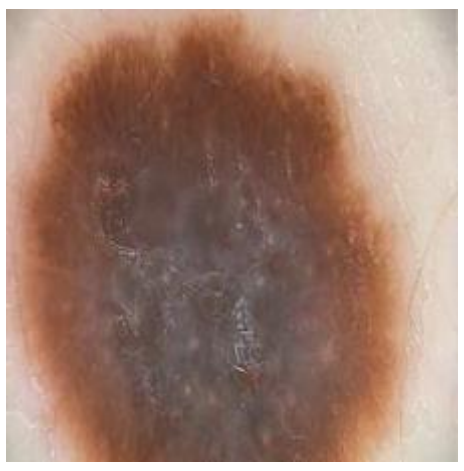 | 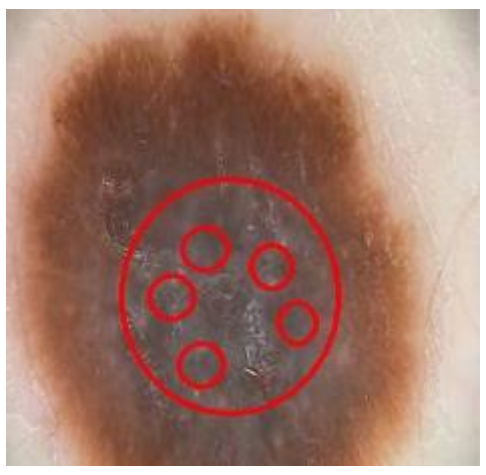 | 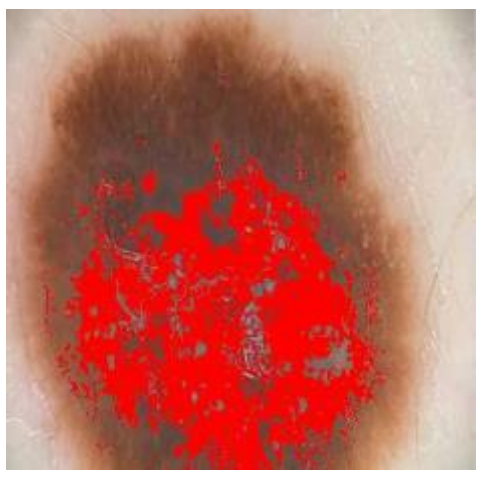 | 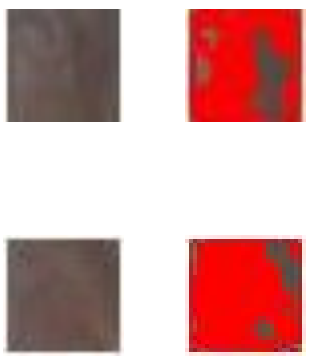 | Red | Min: 86 Max: 123 |
| | | | | | Green | Min: 72 Max: 86 |
| | | | | | Blue | Min: 71 Max: 88 |
| ISIC archive ISIC_0072393 | 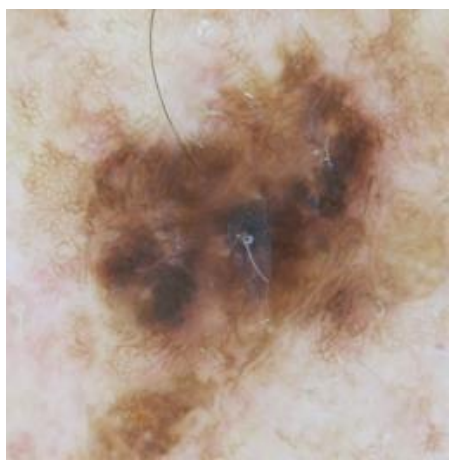 | 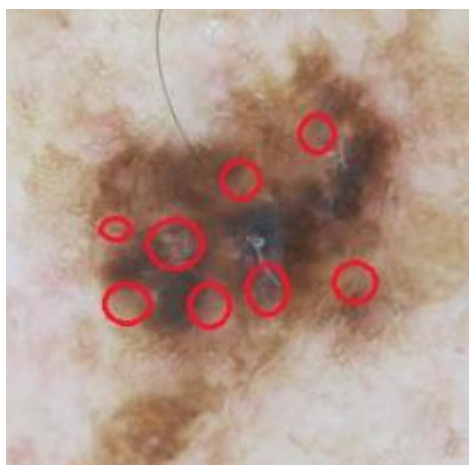 | 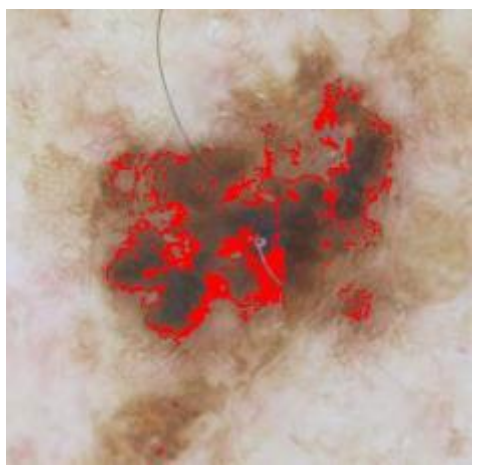 | 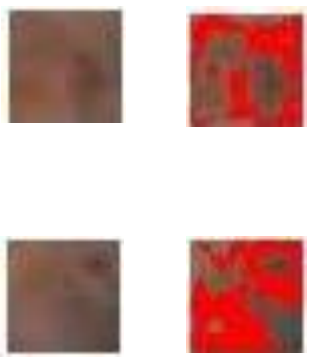 | Red | Min: 86 Max: 135 |
| | | | | | Green | Min: 76 Max: 98 |
| | | | | | Blue | Min: 74 Max: 98 |

In **Table 3**, the first-row image is from the PH2 dataset, and the second-row image is from the ISIC archive dataset. Here, the last column "RGB Value" presents the minimum and maximum value of RGB channels for these two original images which is in the range of BWV colors.

Employing this technique, ground truth is established for 204 dermoscopic images within the ISIC archive dataset, comprising 66 BWV and 138 non-BWV lesions. Since all dermoscopic images of the ISIC archive were captured under constant conditions, there is no requirement for a preprocessing technique. In **Table 4**, Evaluation against dermatologists' decisions revealed the reliability of the proposed imaging technique, serving as a robust method for data labeling in non-annotated datasets. This labeled dataset can be instrumental in training a deep learning model, constituting the second phase of this BWV detection research.

**Table 4**
A comparison between proposed imaging-based BWV detection technique and dermatologists' decision.

| Total=204 | | Ground Truth (Clinical Persons' Opinion) | |
|---|---|---|---|
| | | BWV | Non-BWV |
| Prediction of Proposed Technique | BWV | 66 | 1 |
| | Non-BWV | 0 | 137 |

In this comparison, equations (1) to (3) are applied and found the overall 99.51% AC, 98.51% PR, and 100% SE for the BWV detection on 204 dermoscopic images.

### *3.4 Data Augmentation*

Those three datasets are engaged to train the DCNN model. However, two of them have less amount of data to train a deep model well. Increasing the amount of data is required for those datasets. Data augmentation can help to increase the amount of data. Data augmentation involves image rotation in various angles, flipping, and zooming. **Table 5** reports the different augmentation techniques in PH2 and ISIC archive datasets with the number of regenerated data. The augmentation process is not applied to the Derm7pt dataset, because this dataset already has a decent amount of image data (enough to train a deep CNN model).

**Table 5**
Presentation of the total number of images in three datasets after the augmentation process.

| Dataset | Original | 90 degrees | 180 degrees | 270 degrees | Flip Horizontal | Flip Vertical | Zoom In | Total |
|---|---|---|---|---|---|---|---|---|
| **PH2** | 200 | 200 | 200 | 200 | 200 | 200 | 100 | 1300 |
| **ISIC archive** | 204 | 204 | 204 | 204 | 204 | 204 | 102 | 1326 |
| **Derm7pt** | 2013 | - | - | - | - | - | - | 2013 |

### *3.5 Proposed Deep Convolutional Neural Network*

The initial stage in designing and training a Deep Convolutional Neural Network (DCNN) to distinguish between two classes is to precisely identify the regions of interest in lesion images that are specifically associated with BWV. Due to the challenges presented by these two classes, conventional state-of-the-art deep learning models may not be sufficiently effective for this task. By carefully identifying and isolating the BWV regions within the images, it is possible to train the DCNN to concentrate on these important features. This results in a more reliable classification model, specifically designed for dermoscopic image analysis, with higher accuracy.

The proposed DCNN (31 layers including input and output layers) is inspired by LeCun et al. ***(1998)*** and Rasel et al. ***(2022)***. This inspiration involves various ablation studies and fine-tuning parameters of the proposed DCNN. The ablation study helps to understand the impact of individual components or features by systematically removing them from these two existing models and observing the resulting changes in performance. Besides, fine-tuning allows the proposed DCNN to adapt to a new dataset or task with less data than would be required for training a model from scratch.

Different neural network parameters, such as the number of layers, activation function, kernel or filter size, the number of filters or neurons, dilation factor, padding, and stride, go through different ablation studies and fine-tuning to design an effective deep learning model for the classification of BWV lesions. **Table 6** describes the proposed DCNN layer by layer with different parameters. These parameters are responsible for estimating the network performance. The performance of DCNN is influenced by various parameters, including the number of layers, the number of filters (feature maps), kernel size, stride, padding, activation function, pooling, dropout, learning rate, batch size, optimizer, weight initialization, and data augmentation. Overall, finding the optimal combination of these parameters often involves experimentation and hyper-parameter tuning, as different tasks and datasets might require specific settings for optimal proposed DCNN performance.

Input image data is used from three datasets which are in different dimensions such as PH2: 765-by-572, ISIC archive: 1024-by-1024, and Derm7pt: 768-by-512. For that reason, the input image should be in RGB format with dimensions 256-by-256. Height and width of the filters or kernels, expressed as a vector [*h w*] of two positive integers, where *h* represents height and *w* represents width. The size of local regions that the neurons in the input connect to depends on filter or kernel size. It specifies a positive integer as the number of filters or kernels. According to this condition, there are exactly as many neurons in the layer as there are input regions in the input. This parameter controls how many channels (feature maps) are delivered as a layer. The term 'dilation factor' in this context refers to the factor for dilated convolution, also called atrous convolution, and specified as a vector [*h w*] of two positive integers, where *h* is the vertical dilation and *w* is the horizontal dilation. Without adding more parameters or computations, using dilated convolutions expands the layer's receptive field (the portion of the input it can view). Based on **Table 6**, in 'same' padding, the amount of padding is dynamically calculated which is required to maintain the output size equal to the input size (width or height) when the stride is 1. If the stride is greater than 1, the output size is determined by dividing the input size by the stride and rounding up to the nearest integer (ceiling function). Padding is equally added to the top, bottom, left, and right sides of the input. When the required vertical padding is an odd value, extra padding is added to the bottom to ensure an equal amount on both sides. Similarly, when the required horizontal padding is odd, extra padding is added to the right to maintain symmetry. The ‘same’ padding mode is commonly used in CNNs to ensure that the output feature map has the same spatial dimensions as the input, which can be beneficial in various applications like image segmentation, where preserving spatial information is crucial for accurate results.

**Table 6**
Step-by-step description of the proposed 31-layer DCNN model.

| No. | Layer Type | Kernel Size | Number of Filters / Neurons | Dilation Factor | Padding | Stride |
|---|---|---|---|---|---|---|
| 1 | Image/Input | - | - | - | - | - |
| 2 | Convolution | 5x5 | 8 | 2x2 | same | 2x2 |
| 3 | Normalization | - | - | - | - | - |
| 4 | Custom | - | 8 | - | - | - |
| 5 | Max Pooling | 5x5 | - | - | same | 2x2 |
| 6 | Convolution | 3x3 | 16 | 3x3 | same | 3x3 |
| 7 | Normalization | - | - | - | - | - |
| 8 | Custom | - | 16 | - | - | - |
| 9 | Max Pooling | 3x3 | - | - | same | 3x3 |
| 10 | Convolution | 5x5 | 32 | 2x2 | same | 2x2 |
| 11 | Normalization | - | - | - | - | - |
| 12 | Custom | - | 32 | - | - | - |
| 13 | Max Pooling | 5x5 | - | - | same | 2x2 |
| 14 | Convolution | 3x3 | 64 | 1x1 | same | 1x1 |
| 15 | Normalization | - | - | - | - | - |
| 16 | Custom | - | 64 | - | - | - |
| 17 | Max Pooling | 3x3 | - | - | same | 1x1 |
| 18 | Convolution | 5x5 | 128 | 2x2 | same | 2x2 |
| 19 | Normalization | - | - | - | - | - |
| 20 | Custom | - | 128 | - | - | - |
| 21 | Max Pooling | 5x5 | - | - | same | 2x2 |
| 22 | Convolution | 3x3 | 256 | 1x1 | same | 1x1 |
| 23 | Normalization | - | - | - | - | - |
| 24 | Custom | - | 256 | - | - | - |
| 25 | Max Pooling | 3x3 | - | - | same | 1x1 |
| 26 | Convolution | 5x5 | 512 | 3x3 | same | 3x3 |
| 27 | Normalization | - | - | - | - | - |
| 28 | Custom | - | 512 | | | |
| 29 | Fully connected | - | - | - | - | - |
| 30 | Softmax | - | - | - | - | - |
| 31 | Classification / Output | - | - | - | - | - |

### *3.6 Defining Custom Deep Learning Layer*

Often a custom deep learning layer is required to be built for the classification or regression problem to get better performance. Custom deep learning layers are required in specific scenarios where standard or pre-defined layers provided by popular deep learning frameworks may not fulfill the specific needs of a particular task or model architecture. There are several reasons why custom layers may be necessary: novel architectures, specialized operations, complex operations, efficiency and optimization, domain-specific requirements, model interpretability, transfer learning, pre-training, experimentation, and research. A special deep learning layer called the Parametric Rectified Linear Unit (PReLU) layer (an activation function), which has a learnable parameter, was created, and used in a convolutional neural network ***(Crnjanski et al., 2021)***.

Rectified Linear Unit (ReLU) is a widely used activation function in neural networks, introducing non-linearity by replacing negative values with zero. Leaky ReLU is a variant that addresses the “dying ReLU” problem by allowing a small, positive slope for negative inputs, introducing a form of continuity

***(Maniatopoulos and Mitianoudis, 2021)***. PReLU extends Leaky ReLU by making the negative slope a learnable parameter during training, enhancing adaptability. While ReLU is simple and computationally efficient, Leaky ReLU and PReLU offer solutions to potential issues like dead neurons and vanishing gradients, with PReLU providing added flexibility at the expense of increased model complexity. Any input value that is less than zero for each channel is multiplied by a scalar that was learned during training as part of the PReLU layer's threshold operation. The PReLU layer applies scaling coefficients $\alpha$ to each channel of the input for values less than zero. During training, the layer picks up on these coefficients to create a learnable parameter. The choice among ReLU, Leaky ReLU, and PReLU functions depends on the specific characteristics of the problem and the network's behavior during training. In this research, Leaky ReLU and PReLU are used individually in the hidden layer and later compared with each other's performance. The mathematic formula of the Leaky ReLU and PReLU functions are respectively-

$$f(x) = \begin{cases} x & if\ x > 0 \\ \alpha x & if\ x \leq 0 \end{cases} \quad (6)$$

$$f(x_i) = \begin{cases} x_i & if\ x_i > 0 \\ a_i x_i & if\ x_i \leq 0 \end{cases} \quad (7)$$

In equation (6), $\alpha$*=0.01* is a constant scale factor. In equation (7), $x_i$ is the input of the nonlinear activation $f$ on channel $i$, and $a_i$ is the coefficient controlling the slope of the negative part. The subscript $i$ in $a_i$ indicates that the nonlinear activation can vary on different channels. **Fig. 7** shows the difference between ReLU, Leaky ReLU, and PReLU in a graph.

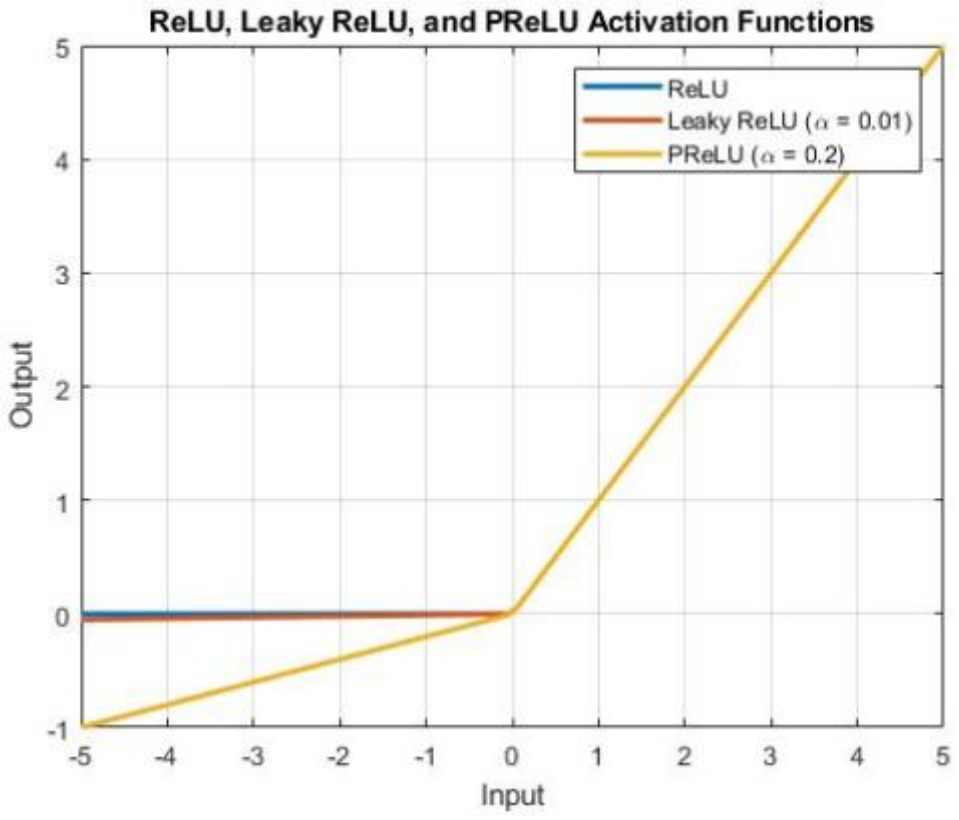

**Fig. 7:** ReLU vs. Leaky ReLU vs. PReLU. For PReLU, the coefficient of the negative part is not constant and is adaptively learned.

### *3.7 Training the Network*

When the number of layers in the proposed DCNN is increased to 29 layers, the challenges of vanishing and exploding gradients arise, making it difficult to train the model effectively. Consequently, both training and testing errors increase significantly. To address this issue, employing techniques such as gradient clipping, normalization, or using appropriate activation functions like ReLU variants is necessary to stabilize the training process and improve model convergence. In this research, two variants of ReLU (Leaky ReLU and PReLU) are used to help mitigate this issue. Additionally, using the stochastic gradient descent with momentum (SGDM) algorithm helps to handle this issue by smoothing gradients, accelerated learning, and improving the stability of the optimization process by reducing oscillations and fluctuations in the parameter updates. Although using Leaky ReLU and PReLU increase the complexity and computational time of the model, they help to improve the model's efficiency and accuracy.

The augmented PH2 (AuP), augmented ISIC archive (AuIa), and Derm7pt datasets are involved separately and collectively to train the proposed model. The image data of each dataset is divided into three sets such as training set, validation set, and test set. The network learnable parameters are updated in a custom training loop using SGDM. The training options are for SGDM, including learning rate (0.01), the maximum number of epochs 250, shuffling the training data before each training epoch and shuffling the validation data before each network validation, frequency of network validation in the number of iterations (25), and displaying plots during network training. The network validation results (using a test set) are shown as confusion matrices to understand the efficiency of the trained network (see **Fig. 8** and **Fig. 9**).

The AuP, AuIa, and Derm7pt datasets are individually and collectively utilized four times to train the proposed DCNN, with each instance considered as an individual experiment. **Table 7** provides details for each experiment, including the training time (TT) for the model. On the right side of the table, the last three columns display the TT for the model using ReLU, Leaky ReLU, and PReLU activation functions, respectively. The model trained with ReLU is included in this table solely to illustrate the difference in computational time.

**Table 7**
The details of all experiments on different datasets with three activation functions and training times.

| Expt. | Training dataset | Training set | Validation set | Number of folds | Maximum Iteration | Testing dataset | TT (ReLU) | TT (Leaky ReLU) | TT (PReLU) |
|---|---|---|---|---|---|---|---|---|---|
| **1** | AuP (100%) | 80% | 20% | 5 | 2250 | ISIC archive (10%) | 233s | 240s | 281s |
| **2** | AuIa (100%) | 80% | 20% | 5 | 2250 | PH2 (10%) | 238s | 245s | 286s |
| **3** | AuP+AuIa (100%) | 80% | 20% | 5 | 4500 | Derm7pt (5%) | 471s | 484s | 567s |
| **4** | Derm7pt (100%) | 80% | 20% | 5 | 3500 | PH2+ISIC archive (10%) | 361s | 371s | 434s |

## 4. Results and Discussion

### *4.1 Experimental Results*

The proposed DCNN is trained on AuP, AuIa, AuP+AuIa, and Derm7pt datasets. At first, Leaky ReLU is used in the custom layer of DCNN as an activation function and trained on all four datasets. Then PReLU is used in the custom layer instead of ReLU and trained again on all four datasets. Training on each dataset considers individual experiments which is a total of 4. That means the proposed DCNN total 8 times is trained in 4 experiments. A confusion matrix (with the first row and column for the BWV class, and the second row and column for the non-BWV class) is generated after each training for different test sets to evaluate the model's performance. **Fig. 8** shows the confusion matrices of the proposed DCNN (Leaky ReLU), and **Fig. 9** shows the confusion matrices of the proposed DCNN (PReLU) from experiments 1 to 4.

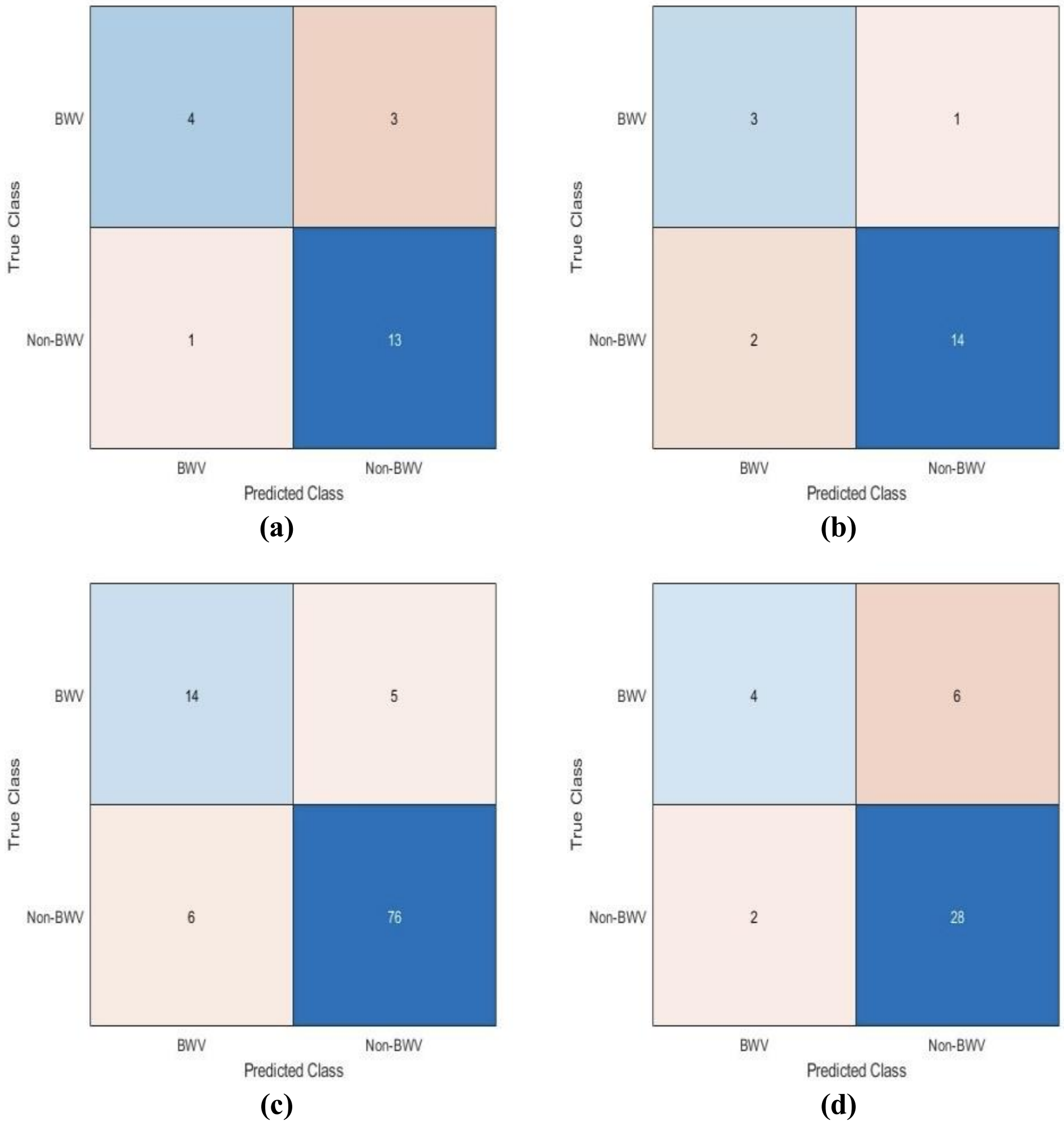

**Fig. 8:** Confusion matrices of the proposed DCNN (Leaky ReLU): (a) training set AuP and testing set 10% of ISIC archive, (b) training set AuIa and testing set 10% of PH2, (c) training set AuP+AuIa and testing set 5% of Derm7pt, and (d) training set Derm7pt and testing set 10% of PH2+ISIC archive.

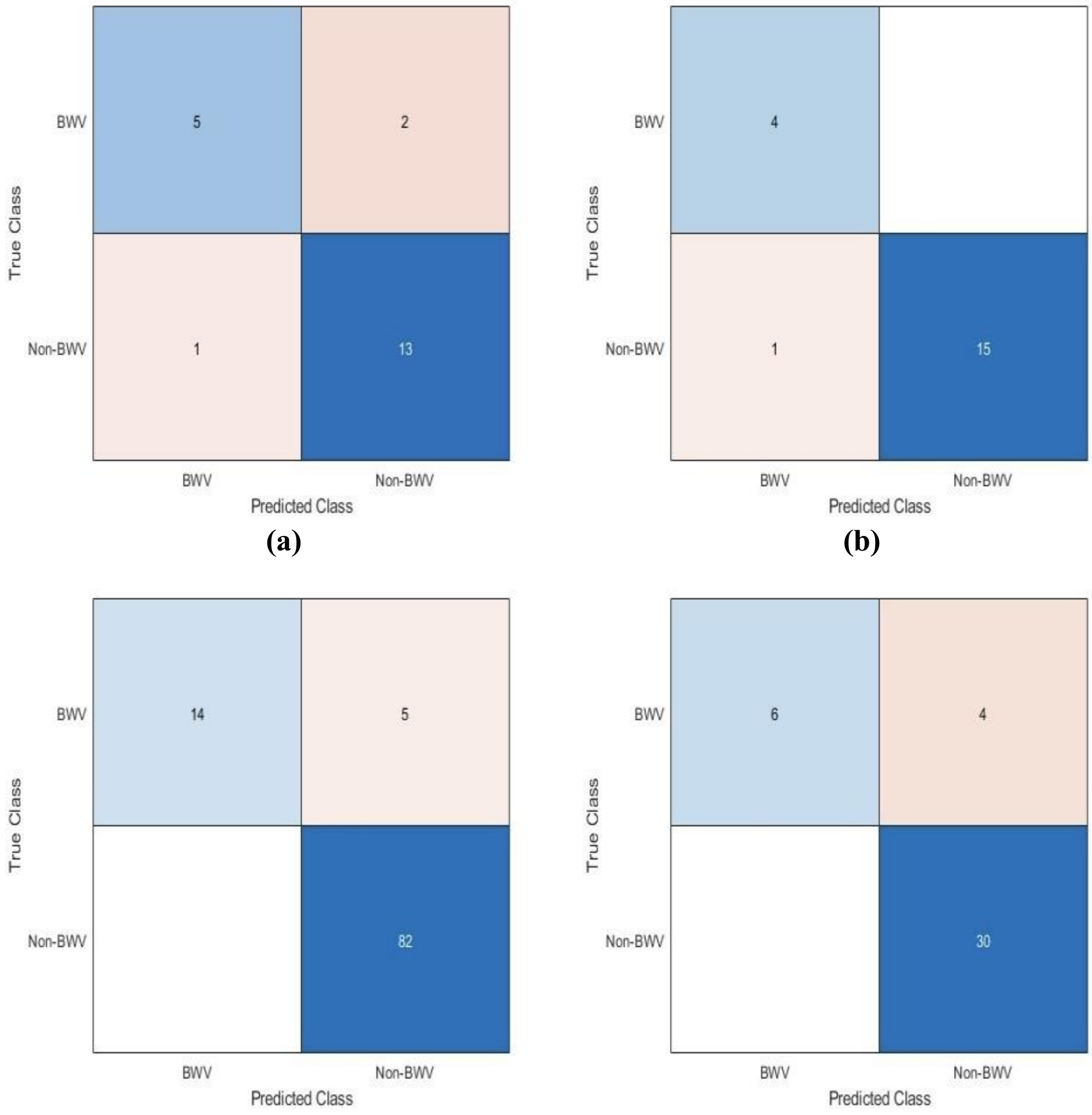

**(a)** **(b)** **(c)** **(d)**

**Fig. 9:** Confusion matrices of the proposed DCNN (PReLU): (a) training set AuP and testing set 10% of ISIC archive, (b) training set AuIa and testing set 10% of PH2, (c) training set AuP+AuIa and testing set 5% of Derm7pt, and (d) training set Derm7pt and testing set 10% of PH2+ISIC archive.

### *4.2 Proposed DCNN vs. Other Approaches*

Equations (1) to (5) are applied to the confusion matrices of **Fig. 8** and **Fig.9**, the result is reported in **Table 8**. From the literature, three different approaches ***(Celebi et al., 2008; Madooei et al., 2019; Madooei and Drew, 2013)*** are implemented for detecting BWV and applied to the same datasets in four experiments. The best effort is made to avoid biases in the classification results. In **Table 8**, the results of the existing three approaches are compared with the achieved results of the proposed DCNN (with Leaky ReLU and PReLU). Moreover, the area under the receiver operating characteristic curve (AUC) is reported for all approaches on all datasets in the same table. To draw the receiver operating characteristic curve, those confusion matrices are used, and the AUC value is calculated.

**Table 8**
The performance comparison between proposed DCNN and other approaches on the four test sets of different datasets.

| Dataset | Approach | AC | PR | SE | F1 | SP | AUC |
|---|---|---|---|---|---|---|---|
| **AuP (Expt. 1)** | Celebi et al. ***(2008)*** | 47.62 | 42.86 | 30.00 | 35.29 | 63.64 | 55.66 |
| | Madooei and Drew ***(2013)*** | 57.14 | 57.14 | 40.00 | 47.06 | 72.73 | 65.00 |
| | Madooei et al. ***(2019)*** | 66.67 | 57.14 | 50.00 | 53.33 | 76.92 | 69.50 |
| | Proposed (Leaky ReLU) | 80.95 | 57.14 | 80.00 | 66.67 | 81.25 | 77.19 |
| | Proposed (PReLU) | **85.71** | **71.43** | **83.33** | **76.92** | **86.67** | **81.75** |
| **AuIa (Expt. 2)** | Celebi et al. ***(2008)*** | 50.00 | 50.00 | 20.00 | 28.57 | 80.00 | 55.40 |
| | Madooei and Drew ***(2013)*** | 60.00 | 75.00 | 30.00 | 42.86 | 90.00 | 64.00 |
| | Madooei et al. ***(2019)*** | 80.00 | 75.00 | 50.00 | 60.00 | 92.86 | 68.60 |
| | Proposed (Leaky ReLU) | 85.00 | 75.00 | 60.00 | 66.67 | 93.33 | 77.20 |
| | Proposed (PReLU) | **95.00** | **100.00** | **80.00** | **88.89** | **100.00** | **89.60** |
| **AuP+AuIa (Expt. 3)** | Celebi et al. ***(2008)*** | 68.32 | 47.37 | 29.03 | 36.00 | 85.71 | 56.10 |
| | Madooei and Drew ***(2013)*** | 80.20 | 52.63 | 47.62 | 50.00 | 88.75 | 67.00 |
| | Madooei et al. ***(2019)*** | 85.15 | 57.89 | 61.11 | 59.46 | 90.36 | 70.70 |
| | Proposed (Leaky ReLU) | 89.11 | 73.68 | 70.00 | 71.79 | 93.83 | 78.30 |
| | Proposed (PReLU) | **95.05** | **73.68** | **100.00** | **84.85** | **94.25** | **89.10** |
| **Derm7pt (Expt. 4)** | Celebi et al. ***(2008)*** | 57.50 | 30.00 | 23.08 | 26.09 | 74.07 | 51.00 |
| | Madooei and Drew ***(2013)*** | 67.50 | 40.00 | 36.36 | 38.10 | 79.31 | 65.60 |
| | Madooei et al. ***(2019)*** | 75.00 | 50.00 | 50.00 | 50.00 | 83.33 | 70.00 |
| | Proposed (Leaky ReLU) | 80.00 | 40.00 | 82.35 | 50.00 | 82.35 | 79.50 |
| | Proposed (PReLU) | **90.00** | **60.00** | **100.00** | **75.00** | **88.24** | **85.50** |

In Experiment 1 (trained on AuP and tested on AuIa), Experiment 2 (trained on AuIa and tested on AuP), Experiment 3 (trained on AuP+AuIa and tested on Derm7pt), and Experiment 4 (trained on Derm7pt and tested on AuP+AuIa), all five models—comprising three existing approaches and two proposed models incorporating Leaky ReLU and PReLU—undergo testing on unfamiliar dermoscopic images. Across all experiments, a consistent performance hierarchy is observed, with PReLU-based DCNN claiming the top position, followed by Leaky ReLU-based DCNN, Madooei et al. in third place, Madooei and Drew in fourth, and Celebi et al. in fifth.

In Experiment 2, the PReLU-based DCNN demonstrates 100% precision and 100% specificity, resulting in a false positive rate of zero. In Experiments 3 and 4, the same PReLU-based DCNN achieves 100% sensitivity, indicating a false negative rate of zero. These slight variations in results are observed due to the use of multiple dermoscopic datasets in the experiments. In contrast, the Leaky ReLU-based DCNN and the other three existing approaches exhibit similar variations in terms of precision, sensitivity, and specificity across all four experiments.

However, the PReLU-based DCNN exhibits an average accuracy, precision, sensitivity, F1-score, specificity, and AUC of 91.44, 76.28, 90.83, 81.42, 92.29, and 86.49, respectively, across the four test sets. Conversely, the Leaky ReLU-based DCNN shows an average accuracy, precision, sensitivity, F1-score, specificity, and AUC of 84.27, 61.46, 73.09, 63.78, 87.69, and 78.05 across the four test sets. This demonstration underscores the superior performance of PReLU layers within the proposed model for detecting BWV lesions.

### *4.3 Understanding the Proposed DCNN's Decision*

The Local Interpretable Model-agnostic Explanations (LIME) technique is employed to understand the decision-making process of the proposed DCNN designed for BWV lesion detection. The model-agnostic approach of LIME and its ability to offer human-understandable explanations is considered a better choice than SHAP for interpreting image classifiers in this study. LIME excels in explaining complex models by approximating them with simpler models, such as regression trees which provide insights into the specific neural network's decision patterns ***(Ribeiro et al., 2016)***.

The proposed DCNN evaluates the relevance of input features, highlighting the importance of specific attributes to the DCNN's decisions. LIME generates feature importance maps by segmenting the input image into features and creating synthetic images with varying feature inclusions. These synthetic images are classified using the proposed DCNN, and a simpler regression model is trained to approximate the DCNN's behavior. Feature importance is computed using this regression model, producing a map that highlights the critical image regions influencing the DCNN's decisions.

At first, we train our proposed DCNN on the AuP, AuIa, AuP+AuIa, and Derm7pt datasets and then use this DCNN to predict BWV on the test sets. To understand the prediction process of our trained model, we apply the LIME technique. Initially, we employ the pre-trained DCNN and extract the image input size and output classes. We evaluate individual images of two skin lesions, IMD031 (from PH2) and ISIC_0072393 (from the ISIC archive), resizing each image to match the network input size. Then, we classify the images to produce classes with the highest classification scores displayed in the image titles. Our DCNN classifies both skin lesion images as BWV class, as shown in **Fig. 10**.

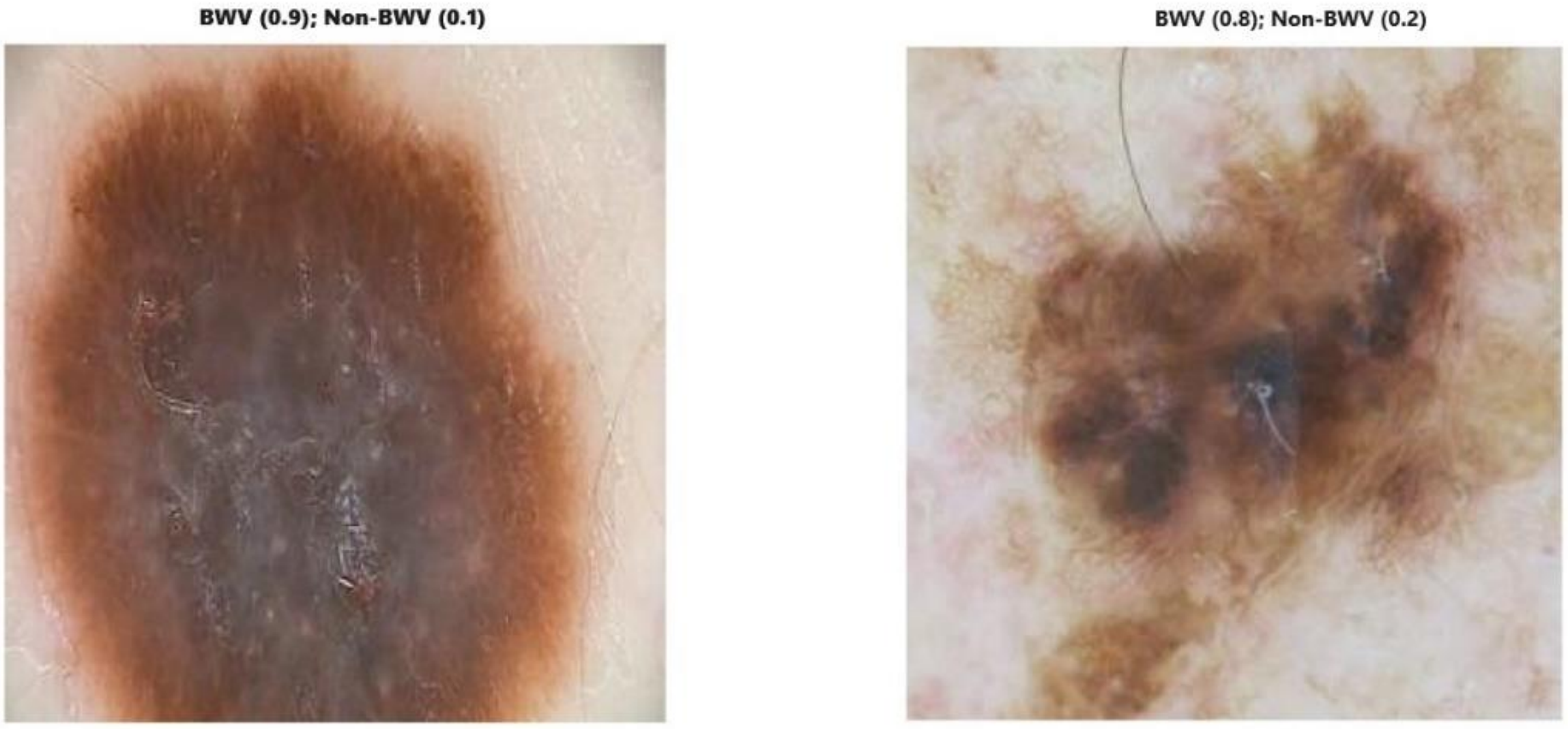

**(a)** **(b)**

**Fig. 10:** The classification decision of DCNN with the LIME calculated probability. Here, **(a)** is IMD031 from PH2 ***(Mendonca et al., 2013)***, and **(b)** is from ISIC_0072393 from ISIC archive: ISIC2016 ***(Gutman et al., 2016)***, ISIC2017 ***(Codella et al., 2017)***, ISIC2018 ***(Codella et al., 2018; Tschandl, 2018)***, and ISIC2019 ***(Tschandl, 2018; Codella et al., 2017; Combalia et al., 2019)***.

Notably, the network also assigns a probability of identifying the skin lesion images as the non-BWV class due to shared features between BWV and non-BWV classes. The influential parts of these images are identified to understand the classification decisions. In both BWV and non-BWV classes, the lesion backgrounds remain consistent. However, when examining the areas of interest within the skin lesion images, we typically identify several common features shared between BWV and non-BWV classes. Nevertheless, distinct features are also present, enabling differentiation between the two classes.

Then, we observe the BWV-predicted class. Several areas of these two images suggest the BWV class. The LIME algorithm segments the input image into superpixels by default to identify different features in the lesion images. The image is segmented into square features (superpixels) using the axes grid lines. To map the significance of various superpixels' properties, the LIME algorithm generates another image (contains different hues over the lesion image). **Fig. 11** shows both BWV images with the LIME maps overlaid. The red hue in the LIME map represents the most important areas or features that influence the model's classification decision. Besides, the blue hue indicates common areas or features that are found in both BWV and non-BWV classes. In the color scale, red hue is the most unique, and blue hue is the most common feature.

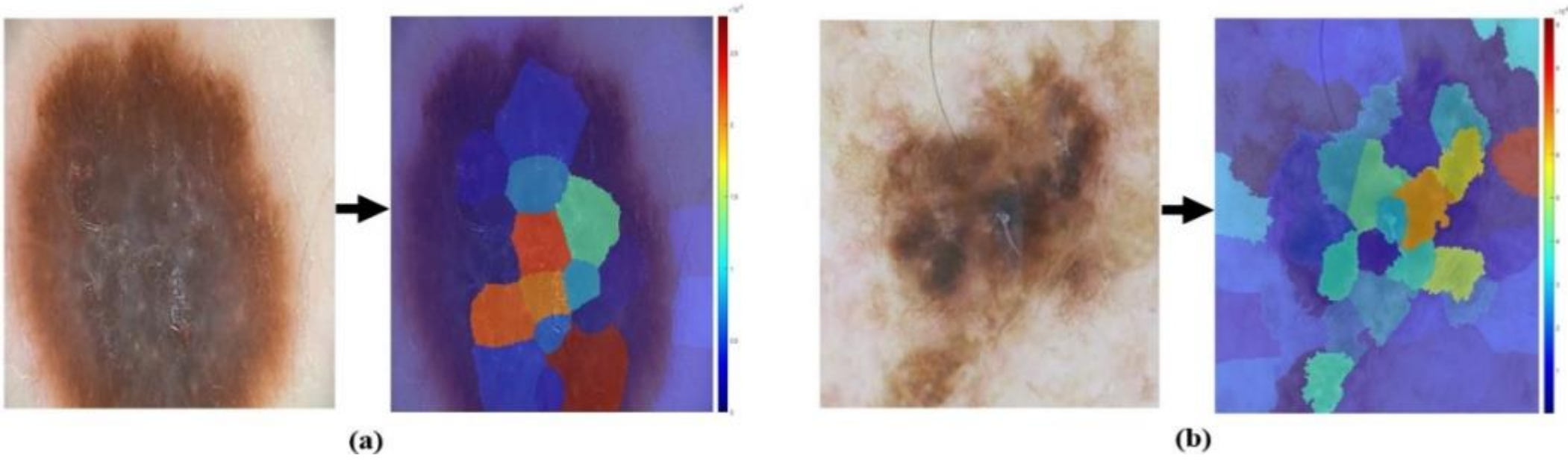

**Fig. 11:** Displaying the LIME calculated important areas of skin lesion images for BWV class.

The LIME maps shown in **Fig. 11** display the portions of the images crucial for BWV classification. The probability score for the BWV class decreases when the red parts of the map are eliminated since they have a higher importance. To construct its BWV forecast, the network concentrates on the red, brown, and bluish regions of the lesions. In the two input images (IMD031 and ISIC_0072393), the proposed model predicted to have the veil with a score of 90% and 80% for the BWV class, and 10% and 20% for the non-BWV class. The highest probability is assigned to the BWV class, with the remaining probability distributed to the non-BWV class when classifying input images. These courses share a portion in common. Comparing the LIME maps calculated for each class reveals which areas of the lesson are more significant for both classes. The non-BWV class's LIME map is generated using the same settings. The red, brown, and light brown areas of the lesion receive more attention from the network than the other areas in the non-BWV class. A visual representation is shown in **Fig. 12**. While both maps highlight the lesions' other colors, the network identifies that the lesions' bluish and whitish area indicate the BWV class. On the other side, the skin lesions' red, brown, and light-brown colors (based on dermatologists' assessment) indicate the non-BWV class. The BWV and non-BWV regions are identified mutually, and they overlap partially.

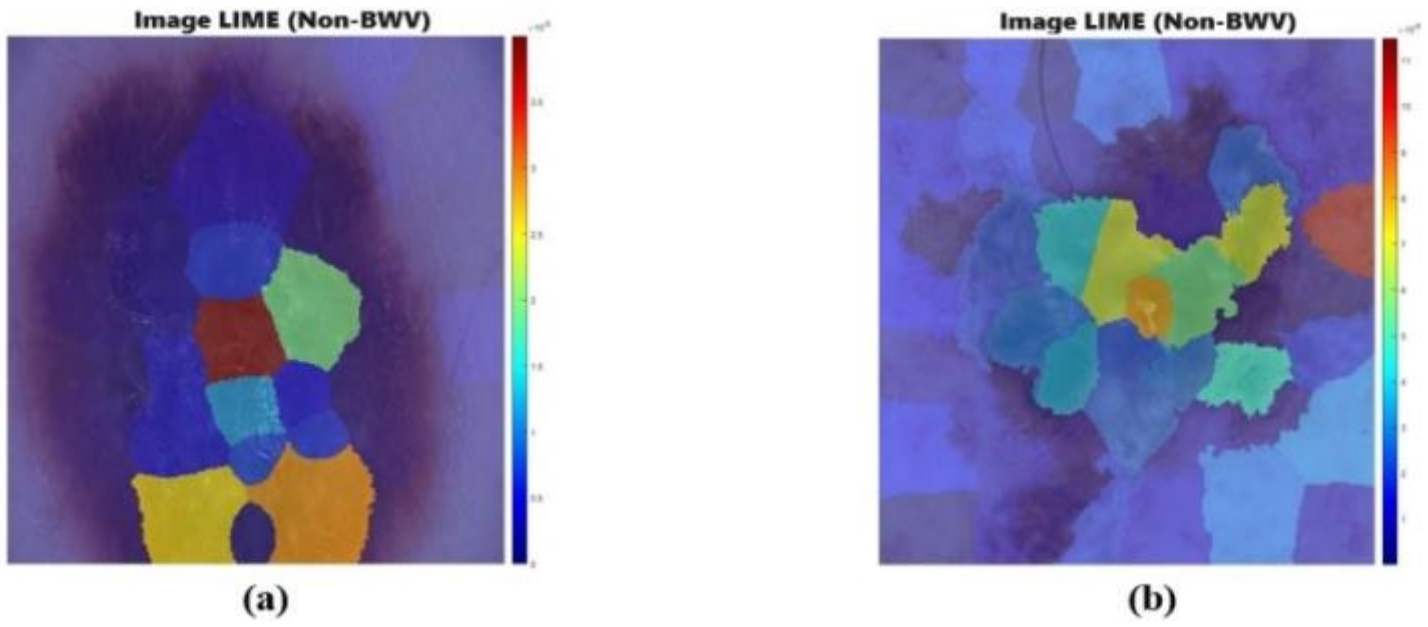

**Fig. 12:** Comparing the LIME calculated non-BWV class maps on the BWV class images to reveal the significant areas for both classes.

To generate a refined heatmap highlighting the lesion image's key areas, we compute the importance of square features and upsample the resulting map, like Grad-CAM. Initially, the image is segmented into a grid of square features rather than irregular superpixels. To align with the image resolution, we upsample the computed map using bicubic interpolation. Additionally, we enhance the resolution of the initially computed map by increasing the number of features to 100, resulting in a 10-by-10 grid of features due to the square shape of the image.

The LIME algorithm creates synthetic images based on the original observation by randomly selecting certain features and replacing all pixels within those features with the average image pixel, effectively eliminating that feature. We increase the number of random samples to 5500, as increasing the number of features yields better results with more samples. We compare the outputs of this process with the previously generated LIME images from **Fig. 11** in **Fig. 13**.

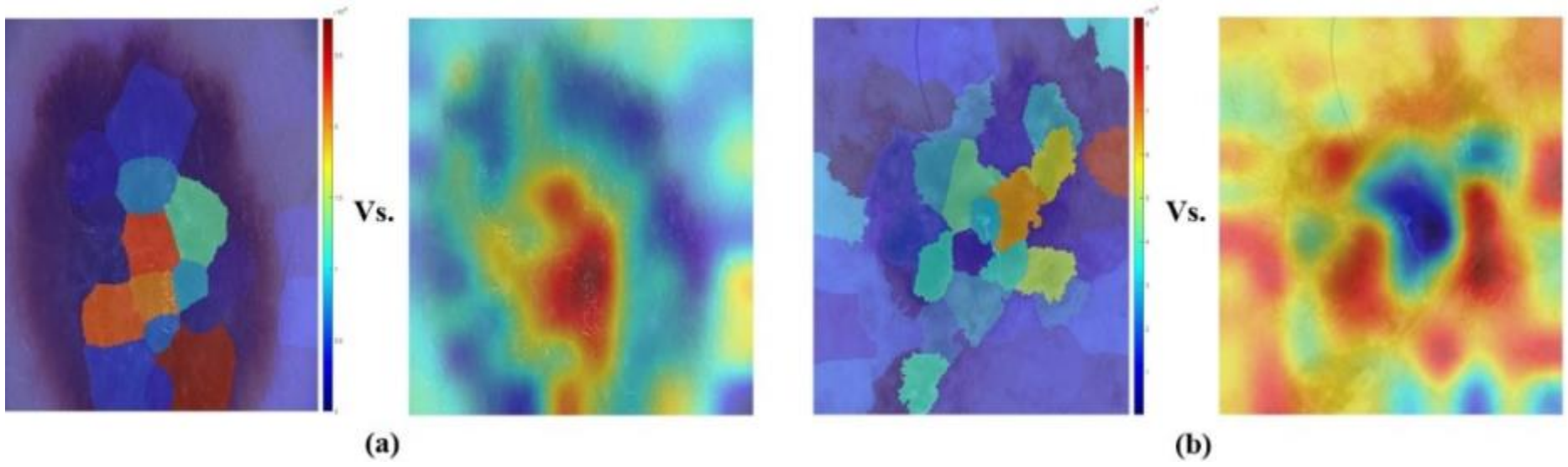

**Fig. 13:** LIME produced smooth heatmaps to highlight the lesion images' key areas for BWV detection.

Then, we obtain the feature maps that highlight the relevance of each input region. From these maps, we identify the most crucial superpixel features and select the top four features for visualization. The LIME-generated maps are used to mask the original images, highlighting only the pixels corresponding to the four most important super-pixels. These masked images, showcasing the most influential features, are presented in **Fig. 14**, building upon the insights provided in **Fig. 10**.

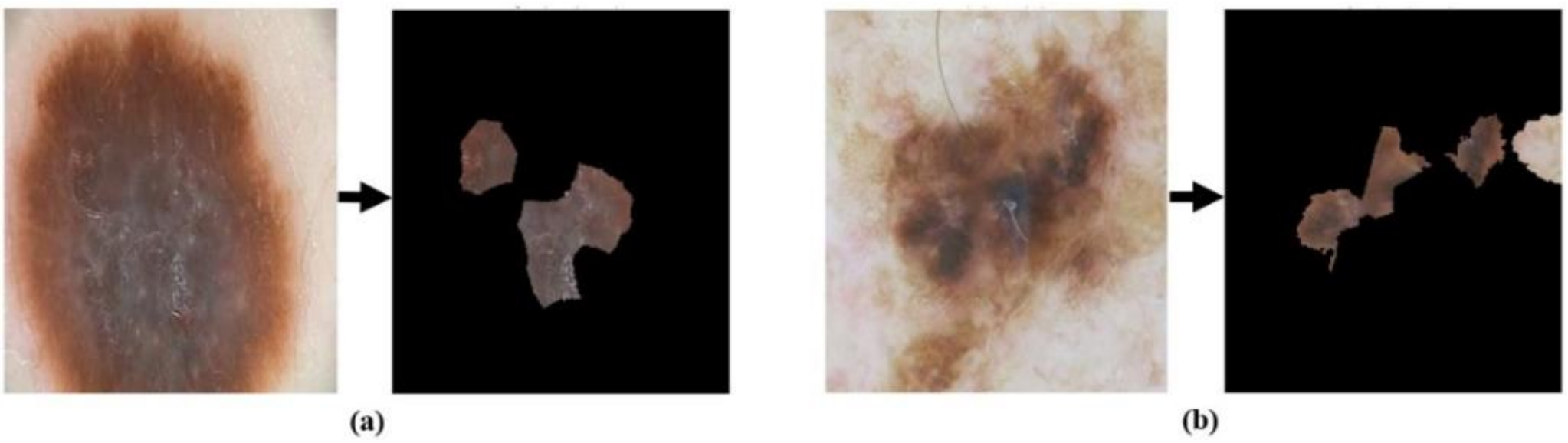

**Fig. 14:** Displaying the most important four features for detecting BWV in two skin lesion images (a) and (b).

## 5. Conclusion and Future Works

In this study, a computer vision algorithm based on image processing was employed to detect the blue-white veil (BWV) on a non-annotated dataset from the ISIC archive, relying on RGB pixel values of lesion patches. Subsequently, a Deep Convolutional Neural Network (DCNN) was crafted, incorporating a custom activation function (PReLU), and trained as a classifier to ascertain the presence or absence of BWV across multiple datasets, including PH2, ISIC archive, and Derm7pt. Additionally, a comparative analysis underscored the efficacy of the proposed DCNN as a classifier, surpassing existing state-of-the-art approaches. Finally, an explainable artificial intelligence (XAI) algorithm was implemented to gain insights into the decision-making process of the DCNN.

The peak performance of the proposed approach on four independent test sets, which were completely unfamiliar to the model, showcased an accuracy of 95.05%, precision of 100.00%, sensitivity of 100.00%, F1-score of 88.89%, specificity of 100.00%, and AUC of 89.10%. These results indicate that the proposed model significantly improves the detection of BWV in skin lesions, providing a robust tool for early melanoma diagnosis.

Future work will involve extending the proposed methodology to identify other prevalent colors in dermoscopic images and validating the efficacy of the identified features in early melanoma detection. The authors are actively developing requisite algorithms and refining additional patterns to construct an integrated system aimed at facilitating the timely detection of melanoma through dermoscopic images. Further validation in real-world clinical settings and continuous refinement of the model will also be pursued to enhance its practical applicability and reliability.

**Conflicts of Interest:** The authors declare that they have no conflicts of interest to report regarding the present study.

## CRediT authorship contribution statement

**M. A. Rasel:** Conceptualization, Methodology, Software, Validation, Writing – Original Draft, Visualization. **Sameem Abdul Kareem**: Conceptualization, Formal analysis, Investigation, Writing - Review & Editing, Supervision. **Zhenli Kwan**: Validation, Formal analysis, Data Curation. **Shin Shen Yong**: Validation, Formal analysis, Data Curation. ***Unaizah H. Obaidellah:*** Conceptualization, Writing-Reviewing and Editing, Supervision, Project administration.

## References

Arroyo, J.L.G., Zapirain, B.G., Zorrilla, A.M., 2011. Blue-white veil and dark-red patch of pigment pattern recognition in dermoscopic images using machine-learning techniques, in: IEEE International Symposium on Signal Processing and Information Technology, ISSPIT 2011. https://doi.org/10.1109/ISSPIT.2011.6151559

Cacciapuoti, S., Di Leo, G., Ferro, M., Liguori, C., Masarà, A., Scalvenzi, M., Sommella, P., Fabbrocini, G., 2020. A measurement software for professional training in early detection of melanoma. Applied Sciences (Switzerland) 10. https://doi.org/10.3390/app10124351

Celebi, M.E., Iyatomi, H., Stoecker, W. V., Moss, R.H., Rabinovitz, H.S., Argenziano, G., Soyer, H.P., 2008. Automatic detection of blue-white veil and related structures in dermoscopy images. Computerized Medical Imaging and Graphics 32. https://doi.org/10.1016/j.compmedimag.2008.08.003

Celebi, M.E., Kingravi, H.A., Aslandogan, Y.A., Stoecker, W. V., 2006. Detection of blue-white veil areas in dermoscopy images using machine learning techniques, in: Medical Imaging 2006: Image Processing. https://doi.org/10.1117/12.655779

Ciudad-Blanco, C., Avilés-Izquierdo, J.A., Lázaro-Ochaita, P., Suárez-Fernández, R., 2014. Dermoscopic findings for the early detection of melanoma: An analysis of 200 cases. Actas Dermosifiliogr 105. https://doi.org/10.1016/j.adengl.2014.07.015

Codella, N. C. F., Gutman, D., Celebi, M. E., Helba, B., Marchetti, M. A., Dusza, S. W., Kalloo, A., Liopyris, K., Mishra, N., Kittler, H., Halpern, A., 2017. Skin lesion analysis toward melanoma detection: A challenge at the 2017 international symposium on biomedical imaging (ISBI), hosted by the international skin imaging collaboration (ISIC). 1710.05006.

Codella, N., Rotemberg, V., Tschandl, P., Celebi, M. E., Dusza, S., Gutman, D., Helba, B., Kalloo, A., Liopyris, K., Marchetti, M., Kittler, H., Halpern, A., 2018. Skin lesion analysis toward melanoma detection 2018: A challenge hosted by the international skin imaging collaboration (ISIC). 1902.03368.

Combalia, M., Codella, N. C. F., Rotemberg, V., Helba, B., Vilaplana, V., Reiter, O., Carrera, C., Barreiro, A., Halpern, A. C., Puig, S., Malvehy, J., 2019. BCN20000: Dermoscopic lesions in the wild. arXiv:1908.02288.

Crnjanski, J., Krstić, M., Totović, A., Pleros, N., Gvozdić, D., 2021. Adaptive sigmoid-like and PReLU activation functions for all-optical perceptron. Opt Lett 46. https://doi.org/10.1364/ol.422930

De Giorgi, V., Massi, D., Trez, E., Salvini, C., Quercioli, E., Carli, P., 2003. Blue hue in the dermoscopy setting: Homogeneous blue pigmentation, gray-blue area, and/or whitish blue veil? Dermatologic Surgery 29. https://doi.org/10.1046/j.1524-4725.2003.29260.x

Di Leo, G., Fabbrocini, G., Paolillo, A., Rescigno, O., Sommella, P., 2009. Towards an automatic diagnosis system for skin lesions: Estimation of blue-whitish veil and regression structures, in: 2009 6th International Multi-Conference on Systems, Signals and Devices, SSD 2009. https://doi.org/10.1109/SSD.2009.4956770

Fabbrocini, G., Betta, G., Di Leo, G., Liguori, C., Paolillo, A., Pietrosanto, A., Sommella, P., Rescigno, O., Cacciapuoti, S., Pastore, F., De Vita, V., Mordente, I., Ayala, F., 2014. Epiluminescence Image Processing for Melanocytic Skin Lesion Diagnosis Based on 7-Point Check-List: A Preliminary Discussion on Three Parameters. Open Dermatol J 4. https://doi.org/10.2174/1874372201004010110

Garrison, Z.R., Hall, C.M., Fey, R.M., Clister, T., Khan, N., Nichols, R., Kulkarni, R.P., 2023. Advances in Early Detection of Melanoma and the Future of At-Home Testing. Life 13, 974. https://doi.org/10.3390/life13040974

Gutman, D., Codella, N. C. F., Celebi, E., Helba, B., Marchetti, M., Mishra, N., Halpern, A., 2016. Skin lesion analysis toward melanoma detection: A challenge at the international symposium on biomedical imaging (ISBI) 2016, hosted by the international skin imaging collaboration (ISIC). 1605.01397.

Kawahara, J., Daneshvar, S., Argenziano, G., Hamarneh, G., 2019. Seven-Point Checklist and Skin Lesion Classification Using Multitask Multimodal Neural Nets. IEEE J Biomed Health Inform 23. https://doi.org/10.1109/JBHI.2018.2824327

Kropidlowski, K., Kociolek, M., Strzelecki, M., Czubinski, D., 2016. Blue whitish veil, atypical vascular pattern and regression structures detection in skin lesions images, in: Lecture Notes in Computer Science (Including Subseries Lecture Notes in Artificial Intelligence and Lecture Notes in Bioinformatics). https://doi.org/10.1007/978-3-319-46418-3_37

Landa, E., Fairchild, M., 2005. Charting Color from the Eye of the Beholder. Am Sci 93. https://doi.org/10.1511/2005.55.975

LeCun, Y., Bottou, L., Bengio, Y., Haffner, P., 1998. Gradient-based learning applied to document recognition. Proceedings of the IEEE 86. https://doi.org/10.1109/5.726791

Madooei, A., Drew, M.S., 2013. A colour palette for automatic detection of blue-white veil, in: Final Program and Proceedings - IS and T/SID Color Imaging Conference. https://doi.org/10.2352/cic.2013.21.1.art00036

Madooei, A., Drew, M.S., Hajimirsadeghi, H., 2019. Learning to Detect Blue-White Structures in Dermoscopy Images With Weak Supervision. IEEE J Biomed Health Inform 23. https://doi.org/10.1109/JBHI.2018.2835405

Manakitsa, N., Maraslidis, G.S., Moysis, L., Fragulis, G.F., 2024. A Review of Machine Learning and Deep Learning for Object Detection, Semantic Segmentation, and Human Action Recognition in Machine and Robotic Vision. Technologies 12, 15. https://doi.org/10.3390/technologies12020015

Maniatopoulos, A., Mitianoudis, N., 2021. Learnable Leaky ReLU (LeLeLU): An Alternative Accuracy-Optimized Activation Function. Information 12, 513. https://doi.org/10.3390/info12120513Henning, J.S., Stein, J.A., Yeung, J., Dusza, S.W., Marghoob, A.A.,

Rabinovitz, H.S., Polsky, D., Kopf, A.W., 2008. CASH algorithm for dermoscopy revisited. Arch Dermatol. https://doi.org/10.1001/archderm.144.4.554

Mendonca, T., Ferreira, P.M., Marques, J.S., Marcal, A.R.S., Rozeira, J., 2013. PH2 - A dermoscopic image database for research and benchmarking, in: Proceedings of the Annual International Conference of the IEEE Engineering in Medicine and Biology Society, EMBS. https://doi.org/10.1109/EMBC.2013.6610779

Olayah, F., Senan, E.M., Ahmed, I.A., Awaji, B., 2023. AI Techniques of Dermoscopy Image Analysis for the Early Detection of Skin Lesions Based on Combined CNN Features. Diagnostics 13, 1314. https://doi.org/10.3390/diagnostics13071314

Rasel, M.A., Obaidellah, U.H., Kareem, S.A., 2022. Convolutional Neural Network-Based Skin Lesion Classification With Variable Nonlinear Activation Functions. IEEE Access 10. https://doi.org/10.1109/ACCESS.2022.3196911

Ribeiro, M.T., Singh, S., Guestrin, C., 2016. "Why should I trust you?" Explaining the predictions of any classifier, in: Proceedings of the ACM SIGKDD International Conference on Knowledge Discovery and Data Mining. https://doi.org/10.1145/2939672.2939778

Seidenari, S., Longo, C., Giusti, F., Pellacani, G., 2006. Clinical selection of melanocytic lesions for dermoscopy decreases the identification of suspicious lesions in comparison with dermoscopy without clinical preselection. British Journal of Dermatology 154. https://doi.org/10.1111/j.1365-2133.2006.07165.x

Soyer, H.P., Argenziano, G., Zalaudek, I., Corona, R., Sera, F., Talamini, R., Barbato, F., Baroni, A., Cicale, L., Di Stefani, A., Farro, P., Rossiello, L., Ruocco, E., Chimenti, S., 2004. Three-Point Checklist of Dermoscopy. Dermatology 208. https://doi.org/10.1159/000075042

Switzer, B., Puzanov, I., Skitzki, J.J., Hamad, L., Ernstoff, M.S., 2022. Managing Metastatic Melanoma in 2022: A Clinical Review. JCO Oncology Practice 18, 335–351. https://doi.org/10.1200/op.21.00686

Tschandl, P., Rosendahl, C., Kittler, H., 2018. The HAM10000 dataset, a large collection of multi-source dermatoscopic images of common pigmented skin lesions. Scientific Data 5. https://doi.org/10.1038/sdata.2018.161

Tsuneki, M., 2022. Deep learning models in medical image analysis. Journal of Oral Biosciences 64, 312–320. https://doi.org/10.1016/j.job.2022.03.003

Wadhawan, T., Hu, R., Zouridakis, G., 2012. Detection of blue-whitish veil in melanoma using color descriptors, in: Proceedings - IEEE-EMBS International Conference on Biomedical and Health Informatics: Global Grand Challenge of Health Informatics, BHI 2012. https://doi.org/10.1109/BHI.2012.6211628

Walter, F.M., Prevost, A.T., Vasconcelos, J., Hall, P.N., Burrows, N.P., Morris, H.C., Kinmonth, A.L., Emery, J.D., 2013. Using the 7-point checklist as a diagnostic aid for pigmented skin lesions in general practice: A diagnostic validation study. British Journal of General Practice 63. https://doi.org/10.3399/bjgp13X667213

Wróblewska-Łuczka, P., Cabaj, J., Bargieł, J., Łuszczki, J.J., 2023. Anticancer effect of terpenes: focus on malignant melanoma. Pharmacological Reports 75, 1115–1125. https://doi.org/10.1007/s43440-023-00512-1